\documentclass[10pt,twocolumn,letterpaper]{article}

\usepackage[pagenumbers]{cvpr}              

\usepackage{array}
\usepackage{algorithm}
\usepackage{algpseudocode}

\usepackage{caption}
\usepackage{multicol}
\usepackage{multirow}

\usepackage{float,wrapfig}

\usepackage{pifont}
\newcommand{\cmark}{\ding{51}}
\newcommand{\xmark}{\ding{55}}

\usepackage{amsmath}
\usepackage{dsfont}

\usepackage{makecell}
\definecolor{cvprblue}{rgb}{0.21,0.49,0.74}
\usepackage[pagebackref=false,breaklinks,colorlinks,allcolors=cvprblue]{hyperref} 
\usepackage{marvosym}
\usepackage{titletoc}

\usepackage{tcolorbox}

\newtcolorbox{codebox}[1][]{
  colback=gray!10,
  colframe=black,
  boxrule=0.4pt,
  arc=2pt,
  top=2pt,
  bottom=2pt,
  left=3pt,
  right=3pt,
  boxsep=3pt,
  fontupper=\small,
  listing only,
  breakable,
  listing options={
    basicstyle=\ttfamily\footnotesize,
    language=Python,
    keywordstyle=\color{blue},
    stringstyle=\color{red},
    commentstyle=\color{gray},
    morekeywords={\quad}
  },
  #1
}

\title{SeeGround: See and Ground for \\Zero-Shot Open-Vocabulary 3D Visual Grounding}

\author{
    Rong Li$^\diamondsuit$\quad Shijie Li$^\triangle$\quad Lingdong Kong$^\heartsuit$\quad Xulei Yang$^\triangle$\quad Junwei Liang$^{\diamondsuit,\square,\textrm{\Letter}}$ \\
    $^\diamondsuit$HKUST(GZ)~~~
    $^\triangle$I$^2$R, A*STAR~~~
    $^\heartsuit$National University of Singapore~~~
    $^\square$CSE, HKUST
}

\begin{document}

\twocolumn[{
\renewcommand\twocolumn[1][]{#1}

\maketitle

\begin{center}
    \centering
    \vspace{-0.5cm}
    \includegraphics[width=\linewidth]{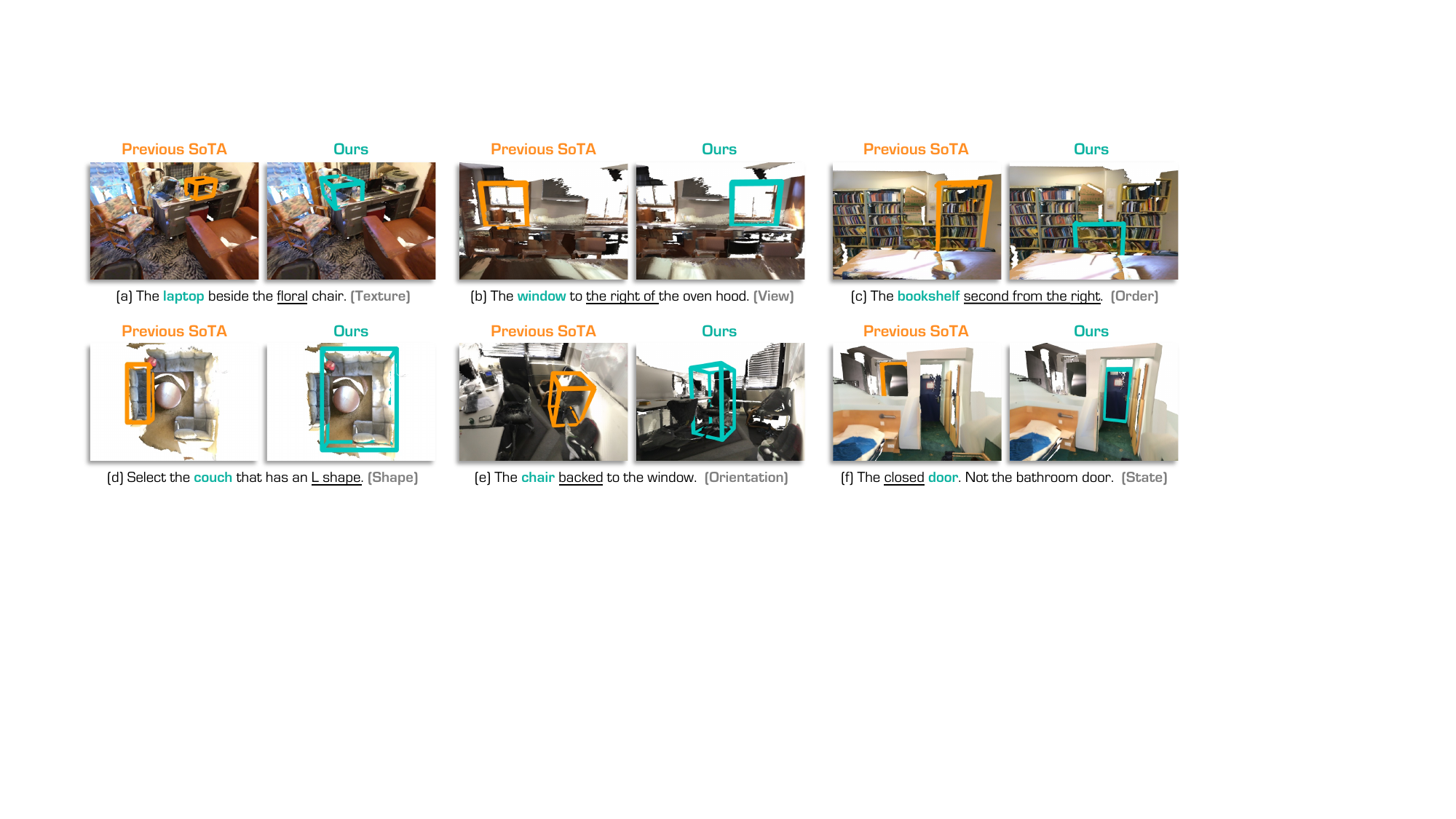}
    \vspace{-0.5cm}
    \captionof{figure}{Effectiveness of \textit{\textbf{\textcolor{BurntOrange}{See}\textcolor{Emerald}{Ground}}}: Different from previous SoTA, our method associates 2D visual cues -- \textit{\textbf{color}}, \textit{\textbf{texture}}, \textit{\textbf{viewpoint}}, \textit{\textbf{spatial position}}, \textit{\textbf{orientation}}, and \textit{\textbf{state}} -- with 3D spatial text description to achieve precise scene understanding. Specifically, our method: (a) identifies the floral chair by recognizing unique color and texture cues; (b) recognizes the couch by interpreting geometric shape; (c) determines the right window by interpreting spatial relationships and perspective; (d) identifies the chair by discerning directional alignment; (e) detects the closed door by visually interpreting its state; and (f) selects the bookshelf by understanding relative positioning.
    } 
    \label{fig:teaser}
\end{center}
}]

\begin{abstract}
3D Visual Grounding (3DVG) aims to locate objects in 3D scenes based on textual descriptions, essential for applications like augmented reality and robotics. 
Traditional 3DVG approaches rely on annotated 3D datasets and predefined object categories, limiting scalability and adaptability. To overcome these limitations, we introduce SeeGround, a zero-shot 3DVG framework leveraging 2D Vision-Language Models (VLMs) trained on large-scale 2D data. SeeGround represents 3D scenes as a hybrid of query-aligned rendered images and spatially enriched text descriptions, bridging the gap between 3D data and 2D-VLMs input formats. We propose two modules: the Perspective Adaptation Module, which dynamically selects viewpoints for query-relevant image rendering, and the Fusion Alignment Module, which integrates 2D images with 3D spatial descriptions to enhance object localization. 
Extensive experiments on ScanRefer and Nr3D demonstrate that our approach outperforms existing zero-shot methods by large margins. Notably, we exceed weakly supervised methods and rival some fully supervised ones, outperforming previous SOTA by 7.7\% on ScanRefer and 7.1\% on Nr3D, showcasing its effectiveness in complex 3DVG tasks. \textbf{Project website (with demo and code): \href{https://seeground.github.io/}{\color{Emerald}\texttt{https://seeground.github.io}}.}
\end{abstract}
\vspace{-0.5cm}    
\section{Introduction}
\label{sec:intro}

3D Visual Grounding (3DVG) aims to locate specific objects within a 3D scene based on given textual descriptions, playing a crucial role in applications such as augmented reality~\cite{chen2020scanrefer, liu2023nips, liu2021tmm, liu2024cvpr, wei2024icme,ma2023examination}, vision-language navigation~\cite{chen2022think,huang2022assister,gong2024cognition}, and robotic 
perception~\cite{chen2023clip2scene,kong2023robo3d,kong2023rethinking,lai2023xvo,li2022coarse3d,zhuang2021perception,tan2024epmf,li2024tfnet,zhuang2024robust, hu2024dhp}. Effective solutions require both textual comprehension and spatial reasoning capabilities within complex 3D environments.

Previous research has focused on specific scenarios, where models~\cite{jain2022bottom, 3dvista, eda, zhao20213dvg, yuan2021instancerefer, chen2020scanrefer, mcln} are trained on small-scale datasets, limiting their scalability and adaptability to diverse, real-world environments. However, gathering large-scale 3D datasets is costly~\cite{behley2019semanticKITTI,sun2020waymoOpen,fong2022panoptic-nuScenes}. Recent studies~\cite{zsvg3d, llmgrounder} attempt to reduce 3D-specific training requirements by reformatting 3D scenes and text descriptions for large language models (LLMs)~\cite{gpt35, openai2023gpt4}, but these methods primarily rely on text input, neglecting rich visual information -- \textit{color}, \textit{texture}, \textit{viewpoint}, \textit{spatial position}, \textit{orientation}, and \textit{state} -- essential for precise localization (\cref{fig:teaser}).

To address these limitations, we propose \textit{\textbf{\textcolor{BurntOrange}{See}\textcolor{Emerald}{Ground}}}, leveraging 2D Vision-Language Models (VLMs)~\cite{openai2023gpt4, qwen2-vl, cogvlm2} for flexible 3DVG. Trained on extensive 2D data, 2D-VLMs offer open-vocabulary understanding ability, performing well in tasks like image captioning \cite{jia2023sceneverse} and visual question answering \cite{3dvista}. This capability provides insight for zero-shot 3DVG. 
Considering that 2D-VLMs cannot process 3D data directly, we introduce a \textbf{cross-modal alignment} representation that enables 2D-VLMs to interpret 3D scenes. This approach combines 2D rendered images with spatially enriched text descriptions, aligning visual and spatial information so that the 2D-VLMs can comprehend the 3D structure and relationships within the scene, thereby achieving more accurate object grounding.

Specifically, we propose to \textbf{represent 3D scenes} as a combination of ``2D rendered images" and ``3D spatial descriptions". The images are rendered using query-driven dynamic viewpoints, simulating relevant observation angles, and capturing object details and spatial context. This approach avoids redundancy in multi-view methods and limitations of bird's-eye views, which lack height and orientation details. The 3D spatial descriptions from pre-saved object detection provide accurate 3D positions, enhancing the VLMs’ understanding of object relationships. 

However, when textual descriptions and images are processed separately by 2D-VLMs, the model cannot associate 3D spatial information from text to the object in the 2D images. For example, in scenes with multiple similar objects (\eg, several chairs), it can be challenging for the model to identify which chair corresponds to the specified one on the image. To address this, we introduce a \textbf{new visual prompting technique} that explicitly marks key objects within images, establishing clear correspondences between 2D images and 3D spatial descriptions. Visual prompting not only enhances the fusion of visual and spatial information but also guides the model’s attention to target areas, reducing potential interference from irrelevant information and improving the localization accuracy in multi-object 3D scenes.

To validate the effectiveness of our approach, we perform extensive experiments on popular benchmarks. We outperform prior zero-shot methods with a $7.7\%$ boost on \textit{ScanRefer} \cite{chen2020scanrefer} and a $7.1\%$ gain on \textit{Nr3D} \cite{achlioptas2020referit3d}, and match some fully supervised ones. Additionally, we conduct a robustness experiment: even with incomplete text input, our method accurately localizes targets by using visual cues from images. Our contributions are as follows:
\begin{itemize}
    \item We introduce \textit{\textbf{\textcolor{BurntOrange}{See}\textcolor{Emerald}{Ground}}}, a training-free solution for zero-shot 3DVG. It converts 3D scenes into 2D-VLMs compatible format
    and combines 2D rendered images with 3D spatial descriptions, leveraging the open-vocabulary capabilities of 2D-VLMs to achieve zero-shot 3DVG without depending on 3D-specific training data.
    \item We design and employ a query-guided viewpoint selection strategy that dynamically adjusts perspectives to capture essential details and spatial context of target objects.
    \item By explicitly associating relative objects in the images to the 3D text description, we establish a specific correspondence between 2D visual features and 3D spatial information, reducing localization ambiguity and boosting efficiency, especially in complex multi-object scenes.
    \item Extensive experiments on the \textit{ScanRefer} and \textit{Nr3D} datasets demonstrate the state-of-the-art performance of our approach across various zero-shot 3DVG tasks.
\end{itemize}
\section{Related work}

\noindent\textbf{3D Visual Grounding.} Supervised 3DVG methods, \eg, ScanRefer \cite{chen2020scanrefer} and ReferIt3D \cite{achlioptas2020referit3d}, achieve object localization by aligning 3D scenes with natural language descriptions. 3DVG-Transformer \cite{zhao20213dvg} further refined localization through attention mechanisms.
More recent work explores enhanced multimodal integration: 
ViewRefer \cite{viewrefer} extends input texts with large language models for multi-view semantic capture. MVT \cite{mvt} and LAR \cite{lar} leverage spatial context from multi-perspective views. SAT \cite{sat} introduces 2D semantic-assisted training to better align 2D-3D features. BUTD-DETR \cite{jain2022bottom} combines bottom-up and top-down detection with transformers. ConcreteNet \cite{concretenet} frames grounding as 3D instance segmentation. WS-3DVG \cite{ws3dvg} uses limited annotations in a coarse-to-fine matching strategy.
PQ3D \cite{pq3d} proposes a unified framework for multiple 3D-VL tasks.
While supervised methods excel on benchmarks, they require extensive annotations, limiting scalability. Recent zero-shot methods, such as LLM-Grounder \cite{llmgrounder} and ZSVG3D \cite{zsvg3d}, are annotation-free and enhance adaptability. However, text-driven approaches often miss critical visual details that could impact precise 3D localization.

\noindent\textbf{3D Open-Vocabulary Understanding.}  
Recent advances in 3D scene understanding enable open-vocabulary capabilities for tasks like segmentation and retrieval through 2D-3D alignments \cite{chen2023clip2scene}. OpenScene \cite{peng2023openscene} achieves open-vocabulary segmentation by projecting 2D pixel-wise features onto 3D scenes. LeRF \cite{kerr2023lerf} integrates multi-scale CLIP features into a neural radiance field.
OVIR-3D \cite{ovir3d} merges multi-view 2D region proposals into 3D space for open-vocabulary instance retrieval.  Agent3D-Zero \cite{agent3d-zero} uses vision-language models across multiple perspectives for 3D question answering and segmentation. RegionPLC \cite{regionplc} generates 3D-text pairs with 2D captions. OpenMask3D \cite{openmask3d} uses aligned images to propose masks for open-vocabulary instance segmentation. OpenIns3D~\cite{openins3d} enhances segmentation across varied 3D scenes through synthetic data.  SAI3D \cite{sai3d} employs Semantic-SAM to acquire 2D masks, connecting them to 3D regions with graph-based merging. These approaches demonstrate the value of 2D-to-3D alignment for open-vocabulary understanding in 3D scenes.

\noindent\textbf{MLLMs Models for 3D Perception.} Recent progress in MLLMs has enhanced 3D understanding by applying 2D techniques to 3D contexts \cite{liu2023seal,li2024is,xu2024superflow}. Scene-LLM \cite{fu2023scene-llm} expands MLLMs' capabilities by supporting 3D dense captioning and semantic segmentation. Uni3DL \cite{li2023uni3dl} introduces a unified framework combining 3D understanding with language comprehension, while 3D-ViSTA \cite{3dvista2023} leverages transformers to align 3D visual data with text inputs, advancing dual-modality comprehension. ConceptFusion \cite{conceptfusion2023} integrates 3D object instances with conceptual knowledge from language, reinforcing 3D semantic understanding and enabling reasoning. Glover~\cite{Ma2024GLOVER} handles 3D manipulation task. SceneVerse \cite{jia2023sceneverse} introduces a language-annotated dataset of 3D environments, helping MLLMs to learn spatial relationships. In addition, RLHF-V \cite{sun2024interactive} enables agents to perform 3D tasks from natural language commands, supporting interactive tasks such as ac and task planning. These models highlight MLLMs' adaptability in enhancing 3D perception, reasoning, and spatial understanding. Our approach builds on these advancements by offering a zero-shot 3D grounding model with improved adaptability, robustness, and broader viewpoint coverage for complex 3D tasks.
\begin{figure*}[t]
    \centering
    \includegraphics[width=\linewidth]{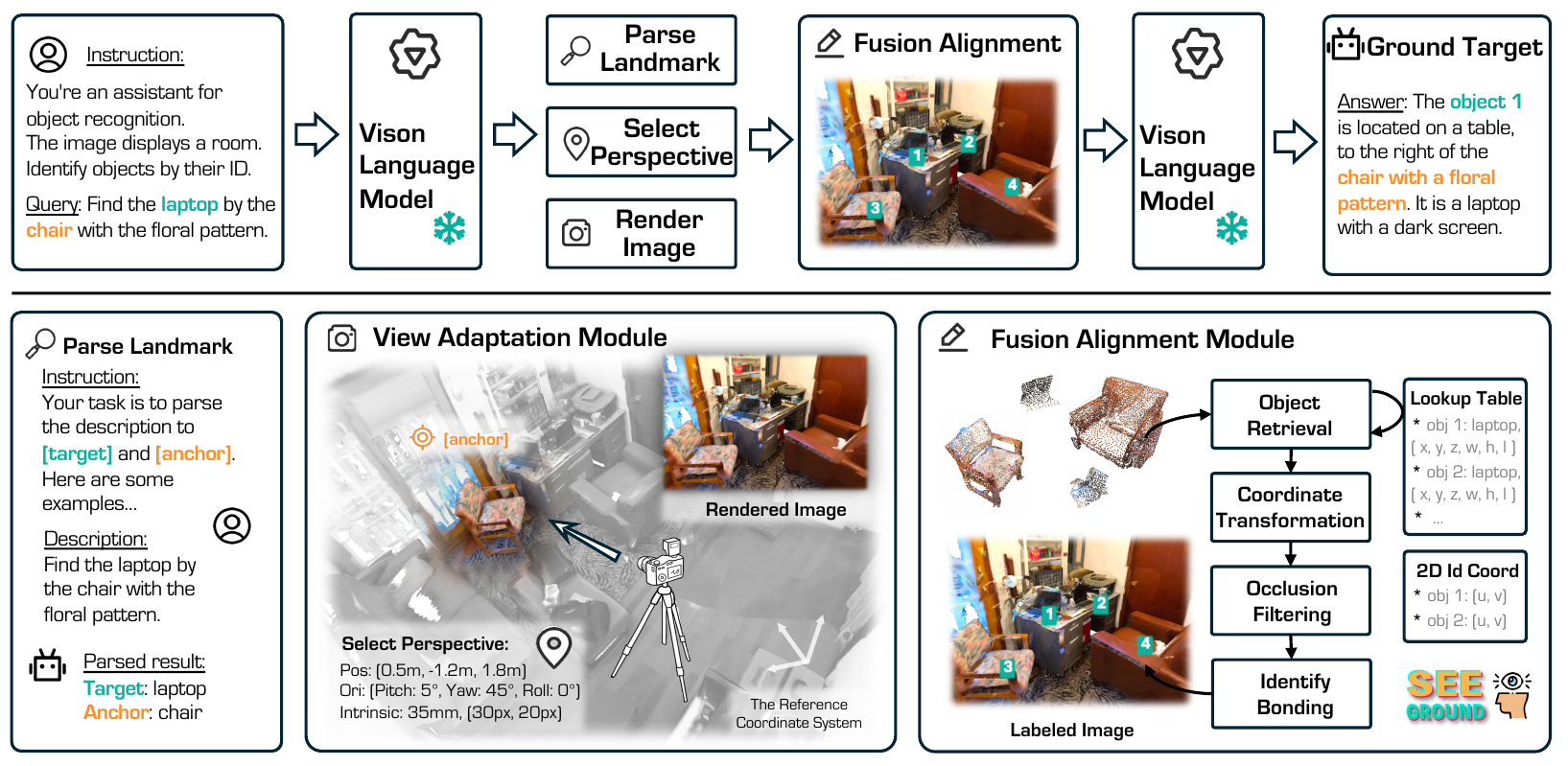}
    \vspace{-0.6cm}
    \caption{Overview of the \textit{\textbf{\textcolor{BurntOrange}{See}\textcolor{Emerald}{Ground}}} framework. We first use a 2D-VLM to interpret the query, identifying both the target object (\eg, ``laptop") and a context-providing anchor (\eg, ``chair with a floral pattern"). A dynamic viewpoint is then selected based on the anchor’s position, enabling the capture of a 2D rendered image that aligns with the query’s spatial requirements. Using the Object Lookup Table ($\mathcal{OLT}$), we retrieve the 3D bounding boxes of relevant objects, project them onto the 2D image, and apply visual prompts to mark visible objects, filtering out occlusions. The image with prompts, along with the spatial descriptions and query, is then input into the 2D-VLM for precise localization of the target object. 
    Finally, the 2D-VLM outputs the target object’s ID, which is then used to retrieve its 3D bounding box from the $\mathcal{OLT}$, providing the final, accurate 3D position in the scene.
    }
    \label{fig:fig2}
     \vspace{-0.2cm}
\end{figure*}

\section{Methodology}
\label{sec:3}

\textbf{Overview. } 
The task of 3DVG aims to precisely locate a target object within a 3D scene $ \mathcal{S} $, based on a textual description $ \mathsf{Q} $. The goal is to output a directed 3D bounding box $(\mathbf{bbox})$ of object $o$ that identifies the target object's location and dimensions. Formally, this process can be expressed as:
\begin{equation}
    \mathbf{bbox} = \mathbf{3DVG}(\mathcal{S}, \mathsf{Q}).
\end{equation}
In this work, we propose a novel method for 3DVG that integrates 2D-VLM with spatially enriched 3D scene representations. Traditional 3D scene models are not directly compatible with the input format required by 2D-VLM, which has shown great promise in scene understanding tasks like image captioning and visual question answering \cite{jia2023sceneverse, qwen2-vl}. To bridge this gap, we introduce a \textbf{hybrid representation}: it utilizes 2D rendered images that are easily processed by 2D-VLM, while also incorporating text-based 3D spatial descriptions. 
This representation allows our framework to align the rich visual features from 2D renderings with the spatial context from 3D scene descriptions. By doing so, we facilitate effective multimodal information alignment and ensure that the 2D-VLM can understand and reason about objects in complex 3D environments without the need for additional 3D-specific training.

We first establish a multimodal 3D representation that is compatible with the 2D-VLM input format in \cref{sec:3.1}, comprising a perspective-aligned rendered image and a spatial description in the text. For each query, the Perspective Adaptation Module (\cref{sec:3.2}) generates a 2D rendered image from a perspective that aligns with the query, capturing relevant objects and spatial relationships. The spatial description, stored in the Object Lookup Table ($\mathcal{OLT}$), includes the 3D bounding boxes and semantic labels of these objects. Further, the Fusion Alignment Module (\cref{sec:3.3}) integrates the rendered image with the 3D spatial descriptions, creating an aligned multimodal representation. This alignment allows the 2D-VLM to process the query, the aligned image, and the text description together, enabling precise localization and retrieval of the target object. The overall framework is illustrated in \cref{fig:fig2}.

\subsection{Multimodal 3D Representation}
\label{sec:3.1}

Our method proposes to leverage 2D-VLM, which is trained on large-scale text-image datasets to capture rich prior knowledge. This enables the model to interpret novel objects and scenes, facilitating open-set understanding. However, prior 3D scene representations -- such as point clouds~\cite{peng2023openscene, hong2023injecting}, voxels~\cite{li2023voxformer}, and implicit representations~\cite{kerr2023lerf} -- are not directly compatible with the input format required by 2D-VLM. To bridge this gap, a 3D scene representation that aligns with 2D-VLM input is necessary.
To tackle this problem, in this work, we propose a hybrid representation that combines ``2D rendered images" and ``text-based 3D spatial descriptions".

\noindent\textbf{Text-based 3D Spatial Descriptions.}
The process begins with detecting all objects in the 3D scene. Using an open-vocabulary 3D detection framework, we identify each object's 3d bounding boxes $\mathbf{bbox}$ (positions and dimensions), and semantic labels $\mathbf{sem}$. This can be formulated as: $(\mathbf{bbox}, \mathbf{sem} )_{i=1}^{N} = \mathbf{OVDet}(\mathcal{S})$.
This information is then converted into a text format compatible with the 2D-VLM input, allowing an accurate spatial and semantic description of the scene.
Since a single scene may correspond to multiple queries, object detection is performed only once per scene, and all $N$ detected objects are stored in the $\mathcal{OLT}$ for efficient processing of subsequent queries. Additionally, the $\mathcal{OLT}$ enables the model to retrieve spatial information efficiently, avoiding complex spatial relationship calculations in later steps.
It is formally defined as follows:
\begin{equation}
    \mathcal{OLT} = \left\{ \left(\mathbf{bbox}, \mathbf{sem} \right)\right\}_{i=1}^N.
\end{equation}

\noindent\textbf{Hybrid 3D Scene Representation.}
While text effectively describes object positions and basic spatial relationships, it often fails to convey critical visual details. To address this, we introduce a multimodal approach that combines image and text descriptions for a more comprehensive 3D scene representation.
Formally, the 3D scene is represented as: 
\begin{equation}
    \left( \mathbf{I}, \mathcal{T} \right) = \mathbf{F}\left( \mathcal{S}, \mathsf{Q}, \mathcal{OLT} \right),
\end{equation}
where $\mathbf{F}$ takes the 3D scene $\mathcal{S}$, the query $\mathsf{Q}$ and the object lookup table  $\mathcal{OLT}$ as inputs, to generate a 2D rendered image  $\mathbf{I}$ and a text-based spatial description $\mathcal{T}$.

The text-based spatial descriptions provide each object's accurate 3D location, dimension, and semantic label, assisting the 2D-VLM in understanding basic spatial relationships between objects. The rendered images offer a 2D perspective of the 3D scene, allowing the model to capture visual features such as color, shape, texture, and relative spatial positions. By combining them, the 2D-VLM gains a comprehensive understanding of both the visual details and true 3D spatial information within the scene.

\subsection{Perspective Adaptation Module}
\label{sec:3.2}
\begin{figure}
    \centering
    \includegraphics[width=\linewidth]{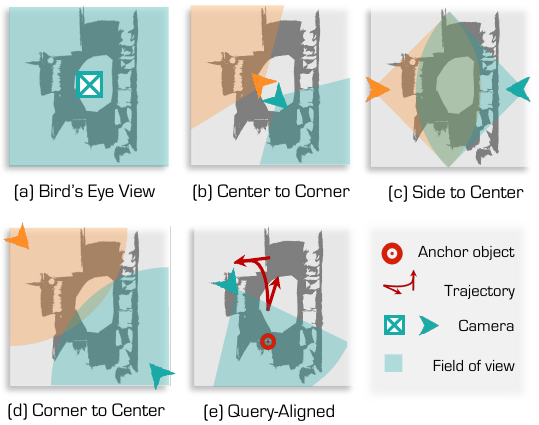}    
    \vspace{-0.6cm}
    \caption{Illustrative example of different perspective selection strategies. Our ``Query-Aligned" method dynamically adapts the viewpoint to match the spatial context of the query, enhancing detail and relevance of visible objects compared to static methods.} 
    \label{fig:view-selection}
    \vspace{-0.2cm}
\end{figure}

There are many strategies for rendering images from the 3D scene, for instance, LAR~\cite{lar} positions the camera around each object to capture multi-view images focused on individual objects. While this provides detailed views, it lacks overall scene context, making it difficult to interpret relationships between objects. Another approach is the bird’s-eye view (BEV), where the camera is positioned above the scene center, capturing a top-down perspective (\cref{fig:view-selection}(a)). This offers a broad scene overview but is limited in height information, causing occlusions in complex 3D environments. To address occlusions, some methods explore multi-view or multi-scale techniques~\cite{openins3d} to capture a wider range of perspectives, as shown in \cref{fig:view-selection} (b)-(d). However, fixed viewpoints often don’t align with the query's perspective, and current 2D-VLM struggles to simulate the speaker’s perspective. Our experiments also reveal that 2D-VLM can misinterpret the scene when rendered images don’t reflect the query’s viewpoint.
To meet these needs, we propose a query-driven dynamic scene rendering method that aligns the rendered viewpoint with the query description, capturing more scene details, as illustrated in~\cref{fig:view-selection} (e).

\noindent\textbf{Dynamic Perspective Selection. }
This process is guided by contextual information in the text instruction $\mathsf{Q}$. The 2D-VLM identifies the anchor object  $ \boldsymbol{A} $ and candidate target objects $\mathcal{O}^{(C)}$ by analyzing relationships in $\mathsf{Q}$. We provide example prompts $\mathcal{E}^{(E)}$ to help the model understand these relationships (see Supp. for prompt construction details). This process is formalized as:
\begin{equation}
  \left( \boldsymbol{A}, \mathcal{O}^{(C)} \right) = \operatorname{VLM} \left( \mathsf{Q}, \mathcal{E}^{(E)} \right).
\end{equation}

Based on the identified anchor $ \boldsymbol{A} $, a rendering viewpoint is selected to capture the scene from a perspective aligned with the query. The viewpoint is initially positioned at the center of the 3D scene, focusing on the chosen anchor $ \boldsymbol{A} $. The camera then moves backward and upward, away from the anchor, to gradually cover a broader view of the scene. 
In the case where there is no anchor object identified during the analysis, \eg, the query describes multiple similar objects), a placeholder anchor is introduced, with the target object serving as the substitute.
If multiple targets are present, the center point of these targets is used as the placeholder anchor. The subsequent perspective selection steps then proceed as described earlier.

\noindent \textbf{Query-Aligned Image Rendering. }
Once the perspective is determined, a virtual camera is placed at this position, and the function $\operatorname{look\_at\_view\_transform}$ (see Supp. for details) generates the rotation matrix $ \mathbf{R}_c $ and translation vector $ \mathbf{T}_c $ based on the virtual camera’s position and orientation relative to the anchor $ \boldsymbol{A} $. 
With these matrices, the scene point cloud is projected and rendered into a 2D image aligned with the query description, represented by:
\begin{equation}
\mathbf{I} = \operatorname{Render}(\mathcal{S}, \mathbf{R}_c, \mathbf{T}_c).
\end{equation}
This approach enables the scene to be observed from a query-consistent viewpoint, providing clear visual details that avoid potential misinterpretations by the 2D-VLM. 
Additionally, as shown in \cref{fig:view-selection} (e), filtering out irrelevant information enhances localization accuracy by reducing interpretive confusion within the model.

\begin{table*}[t]
    \centering
    \caption{Evaluations of 3DVG on \textit{ScanRefer}~\cite{chen2020scanrefer} validation set. Results are reported for \textit{``Unique"} (scenes with a single target object) and \textit{``Multiple"} (scenes with distractors of the same class) subsets, along with overall performance. * indicates results on selected 250 samples.}
    \vspace{-0.2cm}
    \resizebox{\linewidth}{!}{
    \begin{tabular}{r|r|c|c|cc|cc|cc}
        \toprule 
        \multirow{2}{*}{\textbf{Method}} & \multirow{2}{*}{\textbf{Venue}} & \multirow{2}{*}{\textbf{Supervision}} & \multirow{2}{*}{\textbf{Agent}} & \multicolumn{2}{c|}{\textbf{Unique}} & \multicolumn{2}{c|}{\textbf{Multiple}} & \multicolumn{2}{c}{\textbf{Overall}} 
        \\
        & & & & \textbf{Acc@$\mathbf{0.25}$} & \textbf{Acc@$\mathbf{0.5}$} & \textbf{Acc@$\mathbf{0.25}$} & \textbf{Acc@$\mathbf{0.5}$} & \textbf{Acc@$\mathbf{0.25}$} & \textbf{Acc@$\mathbf{0.5}$} 
        \\
        \midrule\midrule
        ScanRefer \cite{chen2020scanrefer} & ECCV'20 & Fully & - & $67.6$ & $46.2$ & $32.1$ & $21.3$ & $39.0$ & $26.1$
        \\
        InstanceRefer \cite{yuan2021instancerefer} & ICCV'21 & Fully & - & $77.5$ & $66.8$ & $31.3$ & $24.8$ & $40.2$ & $32.9$ 
        \\
        3DVG-T \cite{zhao20213dvg} & ICCV'21 & Fully & - & $77.2$ & $58.5$ & $38.4$ & $28.7$ & $45.9$ & $34.5$ 
        \\
        BUTD-DETR \cite{jain2022bottom} & ECCV'22 & Fully & - & $84.2$ & $66.3$ & $46.6$ & $35.1$ & $52.2$ & $39.8$
        \\
        EDA \cite{eda} & CVPR'23 & Fully & - & $85.8$ & $68.6$ &  $49.1$  & $37.6$  & $54.6$ &  $42.3$
        \\
        3D-VisTA \cite{3dvista} & ICCV'23 & Fully & - & $81.6$ & $75.1$ & $43.7$ & $39.1$  & $50.6$ &   $45.8$ 
        \\
        G3-LQ \cite{g3lq} & CVPR'24 & Fully & - & $88.6$ & $73.3$ & $50.2$ & $39.7$  & $56.0$ &   $44.7$ 
        \\
        MCLN \cite{mcln} & ECCV'24 & Fully & - & $86.9$ & $72.7$ & $52.0$ & $40.8$  & $57.2$ &   $45.7$ 
        \\
        ConcreteNet \cite{concretenet} & ECCV'24 & Fully & - & $86.4$ & $82.1$ & $42.4$ & $38.4$  & $50.6$ &   $46.5$ 
        \\
        \midrule
        WS-3DVG \cite{ws3dvg}  & ICCV'23 & Weakly & - & - & - & - & -  & $27.4$ &   $22.0$ 
        \\
        \midrule
        LERF \cite{kerr2023lerf} & ICCV'23 & Zero-Shot & CLIP \cite{clip} & - & -  & - & - & $4.8$ & $0.9$ 
        \\
        OpenScene \cite{peng2023openscene} & CVPR'23 & Zero-Shot & CLIP \cite{clip} & $20.1$ & $13.1$ & $11.1$ & $4.4$ & $13.2$ & $6.5$ 
        \\
        LLM-G \cite{llmgrounder} & ICRA'24 & Zero-Shot & GPT-3.5 \cite{gpt35} & - & -  & - & - & $14.3$ & $4.7$ 
        \\
        LLM-G \cite{llmgrounder} & ICRA'24 & Zero-Shot & GPT-4 turbo \cite{openai2023gpt4} & - & -  & - & - & $17.1$ & $5.3$ 
        \\
        ZSVG3D \cite{zsvg3d} & CVPR'24 & Zero-Shot & GPT-4 turbo \cite{openai2023gpt4} & $63.8$ & $58.4$ & $27.7$ & $24.6$ & $36.4$ & $32.7$
        \\
        VLM-Grounder* \cite{vlmgrounder} & CoRL'24 & Zero-Shot & GPT-4V \cite{openai2023gpt4} & $66.0$ & $29.8$ & $\mathbf{48.3}$ & $\mathbf{33.5}$ & $\mathbf{51.6}$ & $32.8$
        \\
        \textbf{\textcolor{BurntOrange}{See}\textcolor{Emerald}{Ground}} & \textbf{Ours} & Zero-Shot & Qwen2-VL-72b \cite{qwen2-vl} & $\mathbf{75.7}$ & $\mathbf{68.9}$ & $34.0$ & $30.0$ & $44.1$ & $\mathbf{39.4}$
        \\
        \bottomrule
    \end{tabular}}
    \label{tab:tab1}
\end{table*}

\subsection{Fusion Alignment Module}
\label{sec:3.3}
Although the 2D rendered images and text-based spatial descriptions provide substantial spatial information for SeeGround, directly inputting text and images without explicit processing can fail to associate 2D visual features with 3D spatial data. For instance, in scenes with multiple similar objects (\eg, several chairs), the model might struggle to link an object in the image with its corresponding description, leading to grounding errors. 
To address this, we introduce the Fusion Alignment Module, which explicitly associates key visual features in the scene with the textual description, ensuring a clear correspondence between the 2D rendered image and the text-based spatial descriptions.

\noindent\textbf{Depth-Aware Visual Prompting.} 
Specifically, after generating the rendered image $ \mathbf{I} $, the object lookup table $\mathcal{OLT}$ retrieves the bounding box of each candidate object and extracts the 3D points belonging to the object. These points are then projected onto the 2D image plane using the precomputed camera parameters $ \mathbf{R}_c $ and $ \mathbf{T}_c $, and visual markers are placed at the projection locations as prompts.

The simplest approach is to place these markers at the center of the projected points. 
However, due to occlusions inherent in the 3D-to-2D projection process, the projected center may correspond to other objects, misleading the model's understanding of the scene. To address this issue, we leverage depth information to resolve occlusion problems.
For each projected point, its depth is compared with the scene’s depth map to determine visibility. 
Only visible points are used to place the visual prompts.
The visibility of an object $ o $ is determined by evaluating the proportion of its projected points that remain visible.

\begin{figure}[t]
    \centering
    \includegraphics[width=\linewidth]{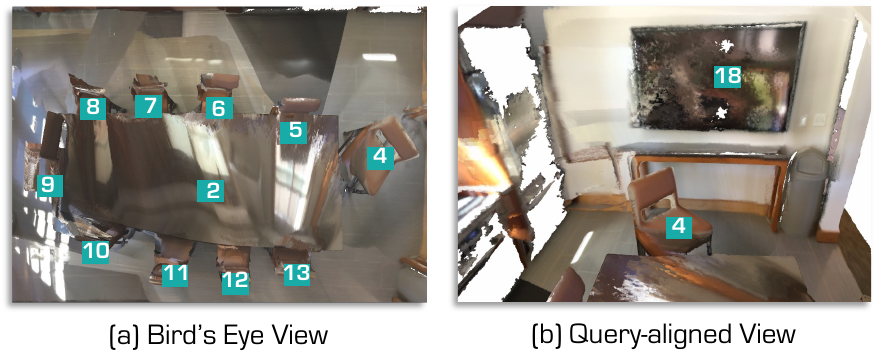}
    \vspace{-0.6cm}
    \caption{Visualization of scene details from different viewpoints. The Bird's Eye View (a) captures the entire scene layout but lacks object-specific detail, while the ``Query-Aligned" View (b) focuses on relevant objects from an optimal angle, revealing additional context like textures and spatial arrangement.}
    \label{fig:exp_projection}
    \vspace{-0.2cm}
\end{figure}

After identifying the visible scene points, we place the visual prompt $ \mathcal{M}_o$ of object $o$ at the projected center of these visible points, ensuring that the prompts reflect the unobstructed positions of the objects. 
The depth-aware visual prompting is expressed as follows:
\begin{equation}
\mathbf{I}_m = \mathbf{I} \odot \left( 1 - \mathds{1}_{\mathcal{P}_{\mathrm{visible}}(o)} \right) + \mathcal{M}_o \odot \mathds{1}_{\mathcal{P}_{\mathrm{visible}}(o)},
\end{equation}
where $ \mathds{1}_{\mathcal{P}_{\mathrm{visible}}(o)} $ is an indicator for the visibility of object $ o $, and element-wise multiplication $ \odot $ is used to apply these prompts selectively. The visualization of a typical example of $\mathbf{I}_m$ is shown in \cref{fig:exp_projection}(b).

\noindent\textbf{Object Prediction with 2D-VLM.}
Finally, given a query $ \mathsf{Q} $, a rendered image $ \mathbf{I}_m $, and the scene's text description $ \mathcal{T} $, the 2D-VLM predicts object $ \hat{o} $ via:
\begin{equation}
    \hat{o} = \operatorname{VLM} \left( \mathsf{Q} \, | \, \mathbf{I}_m, \mathcal{T} \right).
    \vspace{-0.2cm}
\end{equation}

By aligning the visual features in the image with the spatial information in the text, the proposed Fusion Alignment Module effectively reduces ambiguity and improves the model’s localization accuracy, especially in complex scenes with multiple similar objects. 
\section{Experiments}
\label{sec:exp}

\begin{table}[t]
    \centering
    \caption{Performance on \textit{Nr3D}~\cite{achlioptas2020referit3d} validation set. Queries are labeled as \textit{``Easy"} (one distractor) or \textit{``Hard"} (multiple distractors), and as \textit{``View-Dependent"} or \textit{``View-Independent"} based on viewpoint requirements for grounding.}
    \vspace{-0.2cm}
    \resizebox{\linewidth}{!}{
    \begin{tabular}{r|cccc|c}
        \toprule
        \textbf{Method} & \textbf{Easy} & \textbf{Hard} & \textbf{Dep.} & \textbf{Indep.} & \textbf{Overall} 
        \\
        \midrule\midrule
        \multicolumn{6}{l}{\textbf{Supervision: Fully Supervised}}
        \\
        ReferIt3DNet \cite{achlioptas2020referit3d} & $43.6$ & $27.9$ & $32.5$ & $37.1$ & $35.6$ \\
        TGNN \cite{huang2021text} & $44.2$ & $30.6$ & $35.8$ & $38.0$ & $37.3$ \\
        InstanceRefer \cite{yuan2021instancerefer} & $46.0$ & $31.8$ & $34.5$ & $41.9$ & $38.8$ \\
        3DVG-T \cite{zhao20213dvg} & $48.5$ & $34.8$ & $34.8$ & $43.7$ & $40.8$ \\
        BUTD-DETR \cite{jain2022bottom} & $60.7$ & $48.4$ & $46.0$ & $58.0$ & $54.6$ \\
        MiKASA \cite{Chang2024MiKASAM}  & $69.7$ & $59.4$ & $65.4$ & $64.0$ & $64.4$ \\
        ViL3DRel \cite{Chen2022vil3dref}  & $70.2$ & $57.4$ & $62.0$ & $64.5$ & $64.4$ \\
        \midrule
        \multicolumn{6}{l}{\textbf{Supervision: Weakly Supervised}} 
        \\
        WS-3DVG \cite{ws3dvg} & $27.3$ & $18.0$ & $21.6$ & $22.9$ & $22.5$ \\
        \midrule
        \multicolumn{6}{l}{\textbf{Supervision: Zero-Shot}} 
        \\
        ZSVG3D \cite{zsvg3d} & $46.5$ & $31.7$ & $36.8$ & $40.0$ & $39.0$ \\
        \textbf{\textcolor{BurntOrange}{See}\textcolor{Emerald}{Ground}} & $\mathbf{54.5}$ & $\mathbf{38.3}$ & $\mathbf{42.3}$ & $\mathbf{48.2}$ & $\mathbf{46.1}$ \\
        \bottomrule
    \end{tabular}
    }
  \vspace{-0.2cm}
    \label{tab:tab2}
\end{table}

\subsection{Experimental Settings}
\label{sec:4.1}
\textbf{Datasets.}
We use two popular benchmark datasets to evaluate our 3DVG approach. \textbf{ScanRefer}~\cite{chen2020scanrefer} provides 51,500 natural language descriptions across 800 ScanNet scenes, each specifying a target object’s spatial context. Queries are categorized as \textit{``Unique''} (single target) or \textit{``Multiple''} (same-class distractors present), requiring fine discrimination.
\textbf{Nr3D}~\cite{achlioptas2020referit3d}, part of ReferIt3D, includes $41,503$ queries, collected via a two-player reference game to enhance description precision. Queries are divided into \textit{``Easy''} (one distractor) and \textit{``Hard''} (multiple distractors) and are labeled as \textit{``View-Dependent''} or \textit{``View-Independent''} based on whether specific viewpoints are required to ground the target.
\textit{ScanRefer} emphasizes direct 3D localization from sparse point clouds~\cite{chen2020scanrefer}, while \textit{Nr3D} offers ground-truth 3D bounding boxes for all objects~\cite{achlioptas2020referit3d}.

\noindent \textbf{Implementation Details.}
Our main experiments utilize the open-source Qwen2-VL-72B~\cite{qwen2-vl} as the VLM. 
Ablation studies are conducted on the \textit{Nr3D} validation set~\cite{achlioptas2020referit3d}.
The camera captures images of the room at a resolution of $1000\times1000$ pixels, with the top $0.3$ m of the scene excluded to account for the closed room setup. We follow the object detection procedure outlined in ZSVG3D~\cite{zsvg3d} for consistency in evaluation and fair comparison. Due to space limits, please refer to our Appendix for additional details.

\subsection{Comparative Study}
\textbf{ScanRefer.}
\cref{tab:tab1} compares methods on the ScanRefer dataset. our method outperforms other zero-shot methods~\cite{llmgrounder, zsvg3d} and the weakly supervised WS-3DVG~\cite{ws3dvg}, achieving competitive results with supervised methods. In the Unique subset, it achieves Acc@0.25 of $75.7\%$ and Acc@0.5 of $68.9\%$, demonstrating strong performance in single-object scenes. In the challenging Multiple subset, with multiple instances of the target class, our approach attains Acc@0.25 of $34.0\%$ and Acc@0.5 of $30.0\%$, indicating its ability to disambiguate similar objects without supervision. While fully supervised methods like MCLN~\cite{mcln} and ConcreteNet~\cite{concretenet} achieve higher accuracy, our proposed SeeGround framework demonstrates competitive zero-shot performance, highlighting its potential for scalable, annotation-free 3D grounding.

\noindent\textbf{Nr3D.}
\cref{tab:tab2} shows the performance of different approaches on the Nr3D dataset, in which the ground-truth instance mask is also provided. 
Our method achieves $46.1\%$ accuracy on Nr3D, which is a $18.2\%$ improvement over the previous zero-shot baseline, ZSVG3D~\cite{zsvg3d} ($39.0\%)$. In the Easy and Hard categories, our method reaches $54.5\%$ and $38.3\%$, showing robustness across varying scene complexities. For View-Dependent and View-Independent queries, it achieves $42.3\%$ and $48.2\%$, handling different perspectives effectively. While supervised methods like BUTD-DETR~\cite{jain2022bottom} reach $54.6\%$, our method shows that zero-shot methods can achieve competitive performance.

\begin{figure}[t]
    \centering
    \includegraphics[width=\linewidth]{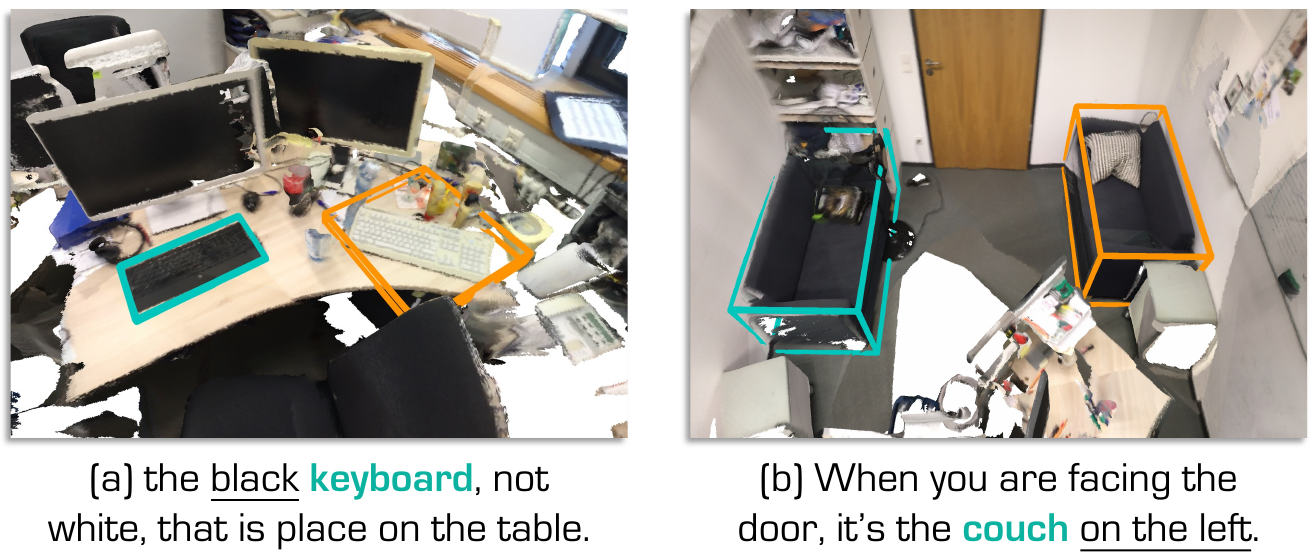}
    \vspace{-0.6cm}
    \caption{\textbf{Qualitative Results.} Rendered images are presented, including the incorrectly identified objects ({\color{BurntOrange}{\textbf{Orange}}}) and correctly identified objects ({\color{Emerald}{\textbf{Green}}}). Key \underline{visual cues} are underlined. 
    }
    \label{fig:vis}
\end{figure}

\begin{table}[t]
    \centering
    \caption{\textbf{Ablation study} on different components in our framework on \textit{Nr3D} \cite{achlioptas2020referit3d}. \textit{``3D Pos."}: 3D object coordinates; \textit{``Layout"}: Scene layout; \textit{``Texture"}: Object color/texture; \textit{``FAM"}: Fusion Alignment Module; and \textit{``PAM"}: Perspective Adaptation Module.}
    \vspace{-0.2cm}
    \resizebox{\linewidth}{!}{
    \begin{tabular}{c|ccccc|c}
    \toprule
    \textbf{\#} & \textbf{3D Pos.} & \textbf{Layout} & \textbf{Texture} & \textbf{FAM} & \textbf{PAM} & \textbf{Overall} 
    \\
    \midrule \midrule
    (a) & \textcolor{Emerald}{\cmark} & \textcolor{BurntOrange}{\xmark} & \textcolor{BurntOrange}{\xmark} & \textcolor{BurntOrange}{\xmark} & \textcolor{BurntOrange}{\xmark} & $37.7$ 
    \\
    (b) & \textcolor{Emerald}{\cmark} & \textcolor{Emerald}{\cmark} & \textcolor{BurntOrange}{\xmark} & \textcolor{BurntOrange}{\xmark} & \textcolor{BurntOrange}{\xmark} & $39.7$ 
    \\
    (c) & \textcolor{Emerald}{\cmark} & \textcolor{BurntOrange}{\xmark} & \textcolor{Emerald}{\cmark} & \textcolor{BurntOrange}{\xmark} & \textcolor{BurntOrange}{\xmark} & $39.5$ 
    \\
    (d) & \textcolor{Emerald}{\cmark} & \textcolor{Emerald}{\cmark} & \textcolor{Emerald}{\cmark} & \textcolor{Emerald}{\cmark} & \textcolor{BurntOrange}{\xmark} & $43.3$ 
    \\
    (e) & \textcolor{BurntOrange}{\xmark} & \textcolor{Emerald}{\cmark} & \textcolor{Emerald}{\cmark} & \textcolor{Emerald}{\cmark} & \textcolor{Emerald}{\cmark} & $45.0$ 
    \\\midrule
    (f) & \textcolor{Emerald}{\cmark} & \textcolor{Emerald}{\cmark} & \textcolor{Emerald}{\cmark} & \textcolor{Emerald}{\cmark} & \textcolor{Emerald}{\cmark} & $\mathbf{46.1}$
    \\
    \bottomrule
\end{tabular}
}
\label{tab:aba}
\end{table}

\subsection{Ablation Study}
\label{sec:4.4}
\noindent\textbf{Effect of Architecture Design.} We begin by evaluating the effectiveness of each module in the proposed architecture. The experimental results are presented in \cref{tab:aba}.

\begin{itemize}
    \item \textit{Layout of the Scene.} Using only 3D coordinates ($37.7\%$, \cref{tab:aba}(a)) provides the basic location of objects but achieves low accuracy. Adding layout ($39.7\%$, \cref{tab:aba}(b)), which renders 3D boxes in 2D without color/texture, improves accuracy by providing spatial context that helps the model understand object positions and sizes.

    \item \textit{Visual Clues.} We find that adding color/texture ($39.5\%$, \cref{tab:aba}~(c)) helps the model distinguish between similar objects, like ``the white keyboard'' versus ``the black keyboard'' (\cref{fig:vis}~(a)). This setup tends to improve accuracy over layout alone by offering object-specific visual cues.

    \item \textit{Fusion Alignment Module.} As shown in \cref{tab:aba}~(d) and \cref{fig:exp_projection}~(b), our proposed Fusion Alignment Module provides a significant increase in accuracy ($43.3\%$) by associating 2D images with text descriptions. 

    \item \textit{Perspective Adaptation Module.} Perspective Adaptation Module ($45.0\%$, \cref{tab:aba}~(e), \cref{fig:exp_projection}~(d)) further improves accuracy by aligning the scene’s viewpoint with the query’s spatial context (\cref{fig:vis}~(b)), helping the model understand the positional context and reducing ambiguity.

    \item \textit{Full Configuration.} We observe that the highest accuracy ($46.1\%$) is achieved with the full configuration (\cref{tab:aba}~(f)). This further validates the effectiveness and efficiency of the proposed SeeGround framework.
\end{itemize}

\begin{figure}[t]
    \centering
    \begin{subfigure}[h]{0.48\linewidth}
        \centering
        \includegraphics[width=\linewidth]{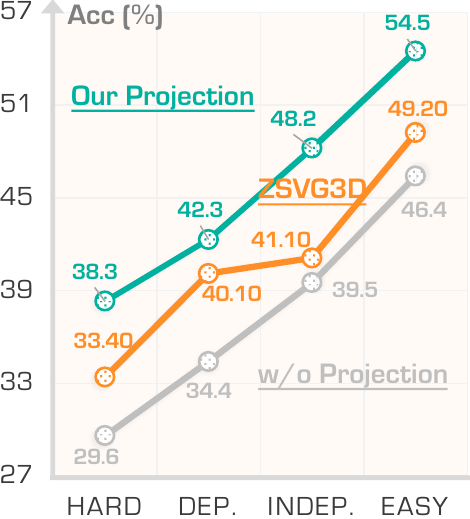}
        \caption{Projection Method}
        \label{fig:compare_projection}
    \end{subfigure}~~
    \begin{subfigure}[h]{0.48\linewidth}
        \centering
        \includegraphics[width=\linewidth]{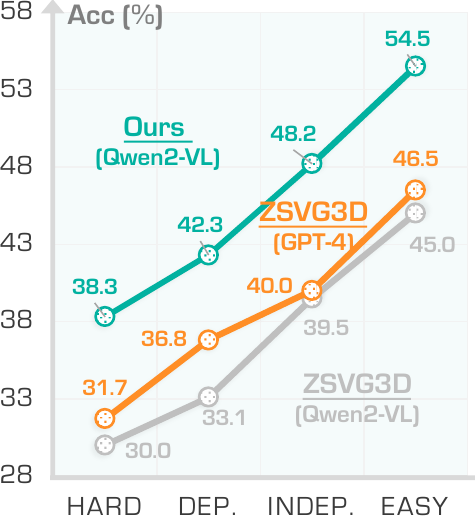}
        \caption{Language Agent}
        \label{fig:compare_agent}
    \end{subfigure}
    \vspace{-0.25cm}
    \caption{\textbf{Ablation study} on using (a) different projection methods (ours \vs ZSVG3D \cite{zsvg3d}); and (b) different language agents (GPT-4 \cite{openai2023gpt4} \vs Qwen2-VL \cite{qwen2-vl}). The results are from \textit{Nr3D} \cite{achlioptas2020referit3d}.}
    \vspace{-0.2cm}
    \label{fig:compare}
\end{figure}

\begin{figure}[t]
    \centering
    \includegraphics[width=\linewidth]{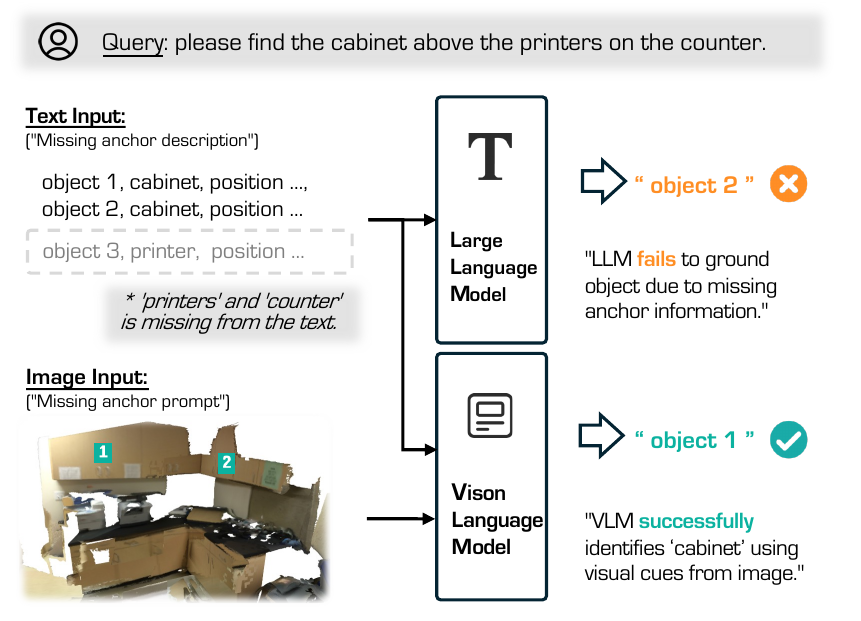}    
    \vspace{-0.7cm}
    \caption{An example of the robustness of the proposed framework in identifying the `cabinet' by leveraging visual context, even when key information (`printers' and `counter') is missing from text input -- an issue that commonly arises in scenarios with detection errors or omissions. Our method is more robust than prior art.}
    \label{fig:robust}
 \vspace{-0.2cm}
\end{figure}

\noindent\textbf{Ours \vs Prior Art.}
ZSVG3D~\cite{zsvg3d} projects object centers onto a 2D image and uses predefined functions to infer spatial relations, but this approach lacks flexibility, omits visual cues, and ignores contextual objects, risking misidentification if detection fails (\cref{fig:robust}). 
\cref{fig:compare_projection} compares the VLM version of ZSVG3D's projection, showing only target and anchor centers, with no background or visual detail. In contrast, our method captures full images, and allows inference of undetected objects via contextual cues \cref{fig:exp_projection}~(b).

\noindent\textbf{Qwen2-VL \vs GPT-4.}
To ensure wider applicability, cost-effectiveness, and reproducibility, we use the open-source model Qwen2-VL~\cite{qwen2-vl} in our method. 
To ensure fairness, we re-evaluate ZSVG3D~\cite{zsvg3d} with Qwen2-VL instead of GPT-4~\cite{openai2023gpt4}, as shown in \cref{fig:compare_agent}, enabling direct comparison with our method. Using the same model, our approach outperforms ZSVG3D across all difficulty levels, confirming its effectiveness independently of model choice. We use ZSVG3D’s program generation prompt with Qwen2-VL, keeping other steps identical.

\begin{table}[t]
    \centering
    \small
    \caption{Performance comparison of different perspective selection strategies. Our method results in consistently higher accuracy across all difficulty levels on Nr3D~\cite{achlioptas2020referit3d} validation set.}
    \vspace{-0.2cm}
    \resizebox{\linewidth}{!}{%
    \begin{tabular}{c|cccc|c}
        \toprule
        \textbf{Type} & \textbf{Easy} & \textbf{Hard} & \textbf{Dep.} & \textbf{Indep.} & \textbf{Overall} 
        \\
        \midrule\midrule
        Center2Corner  & $49.5$ & $31.4$ & $35.1$ & $42.9$  & $40.2$ \\
        Edege2Center   & $51.0$ & $32.7$ & $36.6$ & $44.2$  & $41.5$ \\
        Corner2Center  & $49.8$  & $33.4$  & $35.5$ & $44.5$   & $41.3$ \\
        Bird's Eye View   & $53.4$ & $33.9$ & $36.9$ & $46.8$  & $43.3$ \\
        Query-aligned & $\mathbf{54.5}$ & $\mathbf{38.3}$ & $\mathbf{42.3}$ & $\mathbf{48.2}$ &  $\mathbf{46.1}$ \\
        \bottomrule
    \end{tabular}%
    }
    \vspace{-0.2cm}
    \label{tab:view-selection}
\end{table}

\noindent \textbf{Effect of View Selection Strategy.}
\cref{tab:view-selection} demonstrates the advantage of our dynamic perspective alignment strategy (\cref{fig:view-selection}) over static ones. Static views -- Center2Corner, Edge2Center, and Corner2Center -- lack flexibility and struggle in complex scenes. Bird’s Eye View, though comprehensive, cannot adjust to the query and misses key spatial details like object orientation and height. 
By dynamically adjusting perspective based on the query, our method shows consistent improvement, particularly in \textit{``Hard''} ($4.4\%$) and \textit{``Dependent''} ($5.7\%$).
This result underscores the importance of a flexible and context-aware view selection strategy in 3D scene understanding.

\noindent \textbf{Robustness Evaluation with Incomplete Textual Descriptions.}
As shown in \cref{fig:robust}, we tested our approach's robustness with incomplete textual information, simulating common misdetection scenarios. By omitting an anchor object from the text while retaining the target, our model uses visual cues to compensate, achieving accurate localization. In contrast, LLM performance degrades without the anchor. These results demonstrate that our method maintains high accuracy with partial text, underscoring the importance of integrating visual and textual data for more reliable 3DVG.

\begin{figure}[t]
    \centering
    \begin{subfigure}[h]{0.45\linewidth}
        \centering
        \includegraphics[width=0.95\linewidth]{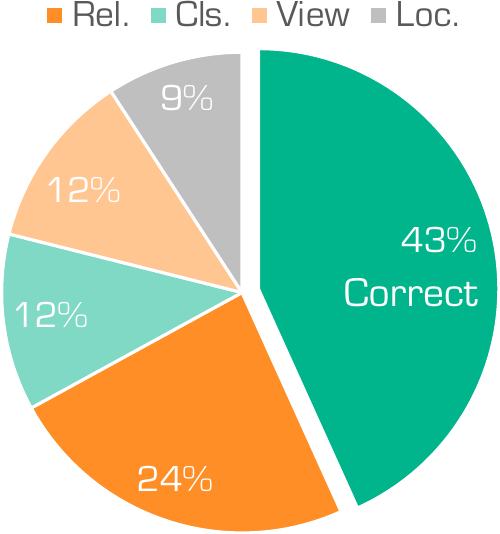}
        \caption{Text-Only Method}
    \end{subfigure}~~~~~~
    \begin{subfigure}[h]{0.45\linewidth}
        \centering
        \includegraphics[width=0.95\linewidth]{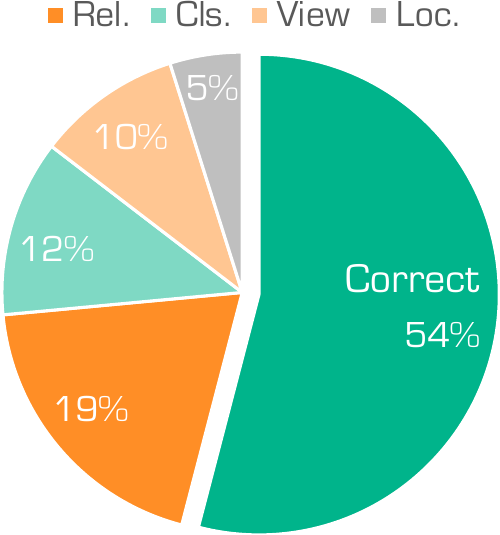}
        \caption{Our Method}
    \end{subfigure}
    \vspace{-0.2cm}
    \caption{Error distributions between the Text-Only Method (a) and Our Method (b), based on four error types: Relation Errors (\textbf{Rel.}, spatial relationship misunderstandings like ``next to" or ``on the corner"), Classification Errors (\textbf{Cls.}, object category misidentifications), Viewpoint Errors (\textbf{View}, errors in interpreting specific observation viewpoints), and Localization Errors (\textbf{Loc.}, errors in pinpointing the target object within the scene). }
    \label{fig:error_source}
    \vspace{-0.2cm}
\end{figure}

\noindent\textbf{Type-Wise Error Analysis.}
To assess the potential limitations of our framework and guide future improvements, we conducted an error analysis on $185$ randomly selected samples across $10$ rooms, manually reviewing predictions to identify key error sources and guide improvements (see~\cref{fig:error_source}). The reduction in spatial localization and target identification errors highlights the importance of visual input in spatial understanding and object recognition. However, despite visual input, our method still has a high error rate in spatial relationships ($19\%$), indicating that precise spatial reasoning remains a challenge. Future work could benefit from advanced spatial reasoning modules. Current viewpoint selection strategies also fall short in handling complex scenarios like ``when the window is on the left'' or ``upon entering from the door''. Finally, high-quality rendering provides clearer information about object boundaries, textures, and colors, helping models more accurately identify and distinguish objects, Our current use of point clouds from the dataset limits rendering precision.
\section{Conclusion}
\label{sec:conclusion}
In this paper, we presented SeeGround, a novel framework for zero-shot 3D Visual Grounding that bridges the gap between 3D data and 2D VLM inputs. By combining query-aligned rendered images and spatial descriptions, SeeGround enables VLMs to interpret 3D scenes without requiring 3D-specific training. Our Perspective Adaptation Module ensures the rendered images align with the query's viewpoint, while the Fusion Alignment Module integrates visual and spatial information, improving localization accuracy and robustness. Experimental results on ScanRefer and Nr3D datasets demonstrate that SeeGround outperforms previous zero-shot methods and achieves competitive performance against supervised models. 
\section*{Acknowledgments} 
This work is supported by the National Natural Science Foundation of China (No. 62306257), and the Guangzhou Municipal Science and Technology Project (No. 2024A03J0619 \& 2024A04J4390). This work is also supported by the LASERi of HKUST(GZ), the Guangzhou-HKUST(GZ) Joint Funding Program (Grant No.2023A03J0008) and Education Bureau of Guangzhou Municipality. This work is also supported by the Agency for Science, Technology and Research (A*STAR) under its MTC Programmatic Funds (Grant No. M23L7b0021). Lingdong Kong is supported by the Apple Scholars in AI/ML Ph.D. Fellowship program.

The authors would like to sincerely thank the Program Chairs, Area Chairs, and Reviewers for the time and efforts devoted during the review process.

\section*{Table of Contents}
\startcontents[appendices]
\printcontents[appendices]{l}{1}{\setcounter{tocdepth}{3}}

\section{Implementation Details}
\label{sec_supp:implement}
In this section, we provide additional details to facilitate the implementation and reproducibility of the proposed \textit{\textbf{\textcolor{BurntOrange}{See}\textcolor{Emerald}{Ground}}} framework.
\subsection{Extended Definitions of Visual Attribute}

In the main body of this paper, we introduce several key visual attributes that are essential for 3D Visual Grounding (3DVG) tasks, including attributes such as texture, shape, viewpoint, and order. \cref{tab:sup-visual-cues} expands upon these attributes, providing more detailed definitions and additional examples to clarify their roles in 3DVG tasks.

This table highlights the indispensable role of visual attributes in disambiguating object references that rely on detailed visual or spatial cues. However, prior approaches \cite{llmgrounder, zsvg3d}, particularly those based on large language models (LLMs), often overlook these attributes due to their reliance on textual inputs alone. Without access to visual information, it becomes challenging for such models to interpret queries like ``\texttt{the black keyboard}'' or ``\texttt{the chair with the tall back}'' This limitation underscores the necessity of incorporating visual information into 3DVG tasks to resolve ambiguities and enrich the alignment between textual queries and 3D spatial contexts.

We hope these analyses could provide insights for future exploration of multimodal systems that integrate textual and visual information in 3DVG.


\begin{table*}[!ht]
    \centering
    \renewcommand{\arraystretch}{1.3}
    \caption{Detailed explanations of \textbf{Key Visual Attributes} in 3D Visual Grounding (3DVG) tasks. In 3DVG, the model must understand the relationships between visual attributes and spatial descriptions in the query to correctly identify and locate the target object. These attributes -- color/texture, shape, viewpoint, order, orientation, state, and functionality -- serve as crucial cues that guide the model.}
    \vspace{-0.2cm}
    \begin{tabular}{>{\raggedright\arraybackslash}m{2.7cm} >{\raggedright\arraybackslash}m{6cm} >{\raggedright\arraybackslash}m{7.5cm}}
    \toprule
    \rowcolor[HTML]{EAEAEA} 
    \textbf{Attribute} & \textbf{Definition} & \textbf{Examples} \\
    \midrule
    \renewcommand{\arraystretch}{1.3}
    \textbf{(a) Texture} & Refers to the \textcolor{BurntOrange}{\textbf{visual appearance}} of an object’s surface, including its color and material properties. These attributes help distinguish objects with similar shapes but different visual properties.
    & 
    \begin{itemize}
        \item ``The \textcolor{BurntOrange}{\texttt{\textbf{black}}} keyboard.''
        \item ``The cart you're looking for is  \textcolor{BurntOrange}{\texttt{\textbf{white}}} on top.''
        \item ``The \textcolor{BurntOrange}{\texttt{\textbf{floral}}} chair.''
        \item ``The correct door has \textcolor{BurntOrange}{\texttt{\textbf{vertical lines}}} on it.''
        \item ``Choose the \textcolor{BurntOrange}{\texttt{\textbf{glass doors}}}.''
        \item ``A brown box with a \textcolor{BurntOrange}{\texttt{\textbf{white label}}} on the front of the box.''
    \end{itemize}
    \\\hline
    \textbf{(b) Shape} & Describes the \textcolor{Emerald}{\textbf{geometric form}} of an object, which allows for differentiation between objects with similar class names and sizes but different geometric structures.
    & 
    \begin{itemize}
        \item ``The \textcolor{Emerald}{\texttt{\textbf{round}}} trash can.''
        \item `` \textcolor{Emerald}{\texttt{\textbf{double door}}}.''
        \item `` \textcolor{Emerald}{\texttt{\textbf{L-shaped}}} couch.''
        \item ``The chair with the \textcolor{Emerald}{\texttt{\textbf{tall back}}}.''
        \item ``Choose the \textcolor{Emerald}{\texttt{\textbf{three seater}}} couch.''
        \item ``The correct couch has a  \textcolor{Emerald}{\texttt{\textbf{90-degree angle}}} bend in it it is not straight.''
    \end{itemize} 
    \\\hline
    \textbf{(c) Viewpoint} & Refers to the \textcolor{BurntOrange}{\textbf{perspective or angle}} from which an object or scene is observed. Viewpoint impacts the visibility and relative positioning of objects.
    & 
    \begin{itemize}
        \item ``When \textcolor{BurntOrange}{\texttt{\textbf{facing}}} the windows, the one on the left.''
        \item ``When \textcolor{BurntOrange}{\texttt{\textbf{entering the room}}}, the sofa is on the right side.''
        \item `` \textcolor{BurntOrange}{\texttt{\textbf{Facing}}} the TV, the chair on the right.''
        \item ``When \textcolor{BurntOrange}{\texttt{\textbf{sitting}}} at the bed, the lamp is on the left corner.''
    \end{itemize} 
    \\\hline
    \textbf{(d) Orientation} &  Describes the \textcolor{Emerald}{\textbf{rotational alignment}} of an object in 3D space, which is essential for distinguishing objects that may appear but are oriented differently.
    & 
    \begin{itemize}
        \item ``The table is \textcolor{Emerald}{\texttt{\textbf{tilted}}} slightly.''
        \item ``the keyboard \textcolor{Emerald}{\texttt{\textbf{at an angle}}}.''
        \item ``The chair whose \textcolor{Emerald}{\texttt{\textbf{back is facing}}} the window.''
        \item ``The picture frame is \textcolor{Emerald}{\texttt{\textbf{leaning}}} at an angle on the shelf.''
    \end{itemize} 
    \\\hline
    \textbf{(e) State} & Refers to the current \textcolor{BurntOrange}{\textbf{condition or status}} of an object (\eg, open/closed, active/inactive), helping to differentiate objects that may appear similar but have different functional states.
    & 
    \begin{itemize}
        \item ``The \textcolor{BurntOrange}{\texttt{\textbf{open}}} door.''
        \item ``The \textcolor{BurntOrange}{\texttt{\textbf{closed}}} book on the desk.''
        \item ``The \textcolor{BurntOrange}{\texttt{\textbf{lit}}} candle on the shelf.''
        \item ``The \textcolor{BurntOrange}{\texttt{\textbf{empty}}} cup on the counter.''
        \item ``The \textcolor{BurntOrange}{\texttt{\textbf{stained}}} carpet in the living room.''
    \end{itemize} 
    \\\hline
    \textbf{(f) Order} & Refers to the relative \textcolor{Emerald}{\textbf{positioning or sequence}} of objects within a scene. This is important for identifying objects in a specific spatial arrangement.
    & 
    \begin{itemize}
        \item ``Of the group of pictures, choose \textcolor{Emerald}{\texttt{\textbf{the second}}} one from the right.''
        \item ``In the row of chairs, pick the \textcolor{Emerald}{\texttt{\textbf{the fourth}}} one from the left.''
        \item ``From the stack of books, take \textcolor{Emerald}{\texttt{\textbf{the third}}} one from the top.''
    \end{itemize} 
    \\\hline
    \textbf{(g) Functionality} & Describes the \textcolor{BurntOrange}{\textbf{intended role or purpose}} of an object in the scene. Functionality helps distinguish between objects with similar appearances but different purposes.
    & 
    \begin{itemize}
        \item ``It is the door to \textcolor{BurntOrange}{\texttt{\textbf{get into the bathroom}}}.''
        \item ``The door \textcolor{BurntOrange}{\texttt{\textbf{for people}}}.''
    \end{itemize} 
    \\
    \bottomrule
    \end{tabular}
    \vspace{-6pt}
    \label{tab:sup-visual-cues}
\end{table*}


\begin{table*}[t]
    \centering
    \caption{An example of the instruction used for prompting the VLMs to identify the target object with a rendered image.}
    \vspace{-0.2cm}
    \renewcommand{\arraystretch}{1.3} 
    \setlength{\tabcolsep}{10pt} 
    \tcbset{
        colback=gray!10, 
        colframe=gray!50, 
        arc=2mm, 
        boxrule=0.3mm, 
        width=\textwidth,
    }
    \begin{tcolorbox}
        \textbf{You are a helpful assistant designed to identify objects based on images and descriptions.} \\
        As shown in the image, this is a rendered image of a room, and the picture reflects your current view. \\
        You should distinguish the target object ID based on your current view. Each object is labeled by a unique number (ID) in red color on its surface. \\\\
        \textbf{Object IDs and their spatial information are as follows:} \\
        \texttt{Object ID}: $1$, \texttt{Type}: cabinet, \texttt{Dimensions}: Width $0.19$, Length $3.21$, Height $0.84$, \texttt{Center Coordinates}: X $1.42$, Y $1.01$, Z $0.41$, \\ 
        \ldots 
        \\\\
        \textbf{The 3D spatial coordinate system is defined as follows:} \\
        The X-axis and Y-axis represent horizontal dimensions, with the Y-axis perpendicular to the X-axis. The Z-axis represents the vertical dimension, with positive values pointing upwards. \\\\
        Please review the provided image and object descriptions, then select the object ID that best matches the given description. Provide a detailed explanation of the features or context that led to your decision. \\\\
        \textbf{Respond in the format:} ``\texttt{Predicted ID: \textless ID\textgreater\ Explanation: \textless explanation\textgreater}'', where \texttt{\textless ID\textgreater} is the object ID and \texttt{\textless explanation\textgreater} is your reasoning. \\\\
        \textbf{The given description is:} ``\texttt{This is the large conference table with many chairs}''.
    \end{tcolorbox}
    \label{tab:instructions}
\end{table*}

\begin{table*}[ht]
    \centering
    \caption{An example of the instruction used for prompting the VLMs to identify the target and anchor objects based on the query.}
    \vspace{-0.2cm}
    \renewcommand{\arraystretch}{1.3} 
    \setlength{\tabcolsep}{10pt} 
    \tcbset{
        colback=gray!10, 
        colframe=gray!50, 
        arc=2mm, 
        boxrule=0.3mm, 
        width=\textwidth,
    }
    \begin{tcolorbox}
        \textbf{You are an assistant designed to identify relationships between objects in a scene.} \\
        Your task is to determine both the target and anchor objects based on the query description provided. \\\\
        \textbf{Here are some examples:}
        \begin{itemize}
            \item ``\texttt{Find the chair that is next to the wooden table}.'' \\
                  \texttt{Target: chair, Anchor: wooden table}
            \item ``\texttt{Identify the lamp that is on top of the desk}.'' \\
                  \texttt{Target: lamp, Anchor: desk}
            \item ``\texttt{Locate the book that is under the coffee table}.'' \\
                  \texttt{Target: book, Anchor: coffee table}
        \end{itemize}
       \ldots 
        \\\\
        Now, based on the query below, provide the names of the target object and the anchor object.\\
        \textbf{Response in the format:}
        ``\texttt{Target: \textless target object\textgreater, Anchor: \textless anchor object\textgreater}''. \\\\
        \textbf{Query:} ``\texttt{Find the bowl that is on the dining table.}''.\\
    \end{tcolorbox}
    \label{tab:supp-anchor-prompt}
\end{table*}

\subsection{Details of Textual Prompt Design}

In this work, we design useful prompts to facilitate the learning of visual attributes for 3DVG.
As illustrated in \cref{tab:instructions}, these prompts include several key components which are summarized as follows:

\begin{itemize}
    \item \textbf{Role Specification:} The prompt begins by defining the assistant's role as an entity designed to identify objects based on images and descriptions. This specification is crucial for setting the context and ensuring that the assistant's actions align with the intended task.

    \item \textbf{Visual Contextualization:} The prompt provides a description of the image, indicating that it is a rendered image of a room. This contextualization helps the assistant to understand the spatial layout and the perspective from which the objects should be identified, which is essential for accurate object recognition.

    \item \textbf{Object Labeling \& Spatial Information:} Each object within the image is labeled with a unique identifier (ID) in red, accompanied by detailed spatial information. This includes object type, dimensions, and center coordinates. Such detailed labeling is vital for distinguishing between objects, especially in complex scenes where multiple objects may have similar appearances.

    \item \textbf{Response Protocol:} The prompt specifies a structured format for the assistant's response, requiring a detailed explanation of the features or context that led to the decision. The response format, ``\texttt{Predicted ID:} \textless \texttt{ID}\textgreater \texttt{Explanation:} \textless \texttt{explanation}\textgreater'', ensures that the assistant's reasoning is transparent and verifiable. This protocol is exemplified by the description, ``\texttt{This is the large conference table with many chairs}'', which serves as a practical application of the identification process.

\end{itemize}

These components are meticulously designed to guide the assistant in leveraging both visual and descriptive information for object identification. The structured format not only ensures clarity and consistency in the assistant's responses but also facilitates effective communication and decision-making. By providing a comprehensive framework, the prompts enable the assistant to perform complex identification tasks with precision and reliability, which is critical in applications requiring high accuracy and interpret-ability.

In the main paper, we also discuss the process of determining anchor and target objects based on the query description. To further clarify this process, we provide an illustrative example of the prompt in \cref{tab:supp-anchor-prompt}. This prompt guides the VLM to identify both the target object and its associated anchor object by analyzing their spatial and semantic relationships as described in the query. This design ensures the model focuses on identifying key objects while maintaining alignment with the query's context.

\subsection{Look-At-View Transform}

In the Perspective Adaptation Module, we utilize the $\operatorname{look\_at\_view\_transform}$ function to compute the extrinsic parameters of the virtual camera. 

Specifically, the camera’s rotation matrix $ \mathbf{R}_c $ and translation vector $ \mathbf{T}_c $ are determined based on the camera’s position $ \mathbf{e} = (x_c, y_c, z_c) $, the anchor point $ \mathbf{at} = (x_a, y_a, z_a) $, and the up vector $ \mathbf{up} $. Below, we provide a formal description of the computation process.
\begin{itemize}
    \item \noindent\textbf{Camera Rotation Matrix $ \mathbf{R}_c $. }  
    The camera rotation matrix $ \mathbf{R}_c \in \mathbb{R}^{3 \times 3} $ aligns the camera's local coordinate system with the world coordinate system. It is derived from the following steps:
    \begin{itemize}
        \item The forward direction ($ \mathbf{z}_{\text{axis}} $) is the normalized vector from the camera to the anchor:
        \begin{equation}
            \mathbf{z}_{\text{axis}} = \frac{\mathbf{at} - \mathbf{e}}{\|\mathbf{at} - \mathbf{e}\|},
        \end{equation}
        where $ \mathbf{at} - \mathbf{e} = (x_a - x_c, y_a - y_c, z_a - z_c) $.

        \item The right direction ($ \mathbf{x}_{\text{axis}} $) is obtained as the normalized cross product of the up vector $ \mathbf{up} $ and $ \mathbf{z}_{\text{axis}} $:
        \begin{equation}
            \mathbf{x}_{\text{axis}} = \frac{\mathbf{up} \times \mathbf{z}_{\text{axis}}}{\|\mathbf{up} \times \mathbf{z}_{\text{axis}}\|}.
        \end{equation}

        \item The up direction ($ \mathbf{y}_{\text{axis}} $) is calculated as the cross product of $ \mathbf{z}_{\text{axis}} $ and $ \mathbf{x}_{\text{axis}} $:
        \begin{equation}
            \mathbf{y}_{\text{axis}} = \mathbf{z}_{\text{axis}} \times \mathbf{x}_{\text{axis}}.
        \end{equation}

        \item Finally, $ \mathbf{R}_c $ is constructed by stacking these three vectors:
        \begin{equation}
            \mathbf{R}_c = 
            \begin{bmatrix}
                \mathbf{x}_{\text{axis}} & \mathbf{y}_{\text{axis}} & \mathbf{z}_{\text{axis}}
            \end{bmatrix}^\top.
        \end{equation}
    \end{itemize}

    \item \noindent\textbf{Camera Translation Vector $ \mathbf{T}_c $. }  
    The translation vector $ \mathbf{T}_c \in \mathbb{R}^3 $ corresponds to the position of the camera in the world coordinate system:
    \begin{equation}
        \mathbf{T}_c = \mathbf{e} = (x_c, y_c, z_c).
    \end{equation}

    The $ \operatorname{look\_at\_view\_transform} $ function provides a systematic way to compute the extrinsic parameters of a camera in 3D space. The rotation matrix $ \mathbf{R}_c $ transforms world coordinates into the camera’s view, while the translation vector $ \mathbf{T}_c $ represents the camera’s position. These parameters are essential for rendering and aligning 3D scenes to match the desired perspective.

    \item \noindent\textbf{Query-Aligned Image Rendering.}
    Once $ \mathbf{R}_c $ and $ \mathbf{T}_c $ are computed, the 3D scene $ \mathcal{S} $ is projected into a 2D image plane to render the query-aligned image $\mathbf{I}$:
    \begin{equation}
        \mathbf{I} = \operatorname{Render}(\mathcal{S}, \mathbf{R}_c, \mathbf{T}_c).
    \end{equation}
    This ensures that the rendered image captures the spatial relationships and visual context described in the query.
\end{itemize}

\subsection{Details of Depth-Aware Visual Prompting}

In our method, depth-aware visual prompting plays a key role in aligning 3D spatial information with 2D visual representations while addressing challenges like occlusion during projection. This section elaborates on the technical details and additional considerations involved in this process, which extends the explanation in the main text.

\begin{itemize}
    \item \noindent\textbf{Generating Visual Prompts.}
    To create visual prompts, we first retrieve the 3D bounding boxes and associated point sets for candidate objects from the OLT. These 3D points are then projected onto the 2D image plane using the camera parameters $ \mathbf{R}_c $ (rotation matrix) and $ \mathbf{T}_c $ (translation vector) obtained during the rendering process. Specifically, for a given 3D point $ p = (x, y, z) $, the 2D projection $ p' = (x', y') $ is computed as:
    \begin{equation}
        p' = \mathbf{R}_c p + \mathbf{T}_c.
    \end{equation}
    Once projected, visual markers are initially placed at the center of the projected points for each object. This provides a basic alignment of the object's location within the rendered image.

    \item \noindent\textbf{Addressing Occlusion Using Depth Information.}
    However, projecting 3D points onto a 2D plane often introduces occlusions, where some parts of an object may overlap with other objects or the background. Directly placing visual prompts without accounting for occlusions can lead to ambiguity and misalignment between the visual markers and the object they represent. To address this, depth information is utilized to determine the visibility of each point.
    For every pixel $ p' $ on the 2D image, the scene’s depth map $ \mathcal{D}(p') $ stores the smallest depth value among all points projected to that pixel. Formally, the depth map is defined as:
    \begin{equation}
        \mathcal{D}(p') = \min_{p_s \in \mathcal{S}} d_s,
    \end{equation}
    where $ \mathcal{S} $ is the set of all 3D points in the scene, and $ d_s $ is the depth of point $ p_s $ relative to the camera. To check the visibility of a 3D point $ p $, its depth $ d_p $ is compared with $ \mathcal{D}(p') $ at its projected location $ p' $. A point is considered visible if:
    \begin{equation}
        \text{Visible}(p) = 
        \begin{cases} 
            1, & \text{if ~} d_p < \mathcal{D}(p')~, \\
            0, & \text{otherwise}~.
        \end{cases}
    \end{equation}

    \item \noindent\textbf{Object-Level Visibility and Prompt Placement.}
    To determine whether an object should be visually prompted, the visibility of its constituent points is aggregated. An object $ o $ is considered visible if a sufficient fraction of its points passes the visibility check:
    \begin{equation}
        \text{Visible}(o) = 
        \begin{cases} 
            1, & \text{if } \sum_{p_o \in \mathcal{P}_o} \text{Visible}(p_o) \geq \alpha \cdot |\mathcal{P}_o|, \\
            0, & \text{otherwise},
        \end{cases}
    \end{equation}
    where $ \mathcal{P}_o $ is the set of 3D points corresponding to object $ o $, $ |\mathcal{P}_o| $ is the total number of points for the object, and $ \alpha $ is a threshold factor determining the minimum fraction of visible points required.
    
    By filtering out occluded points and using only visible points to place visual markers, the depth-aware prompting process ensures accurate alignment of 2D visual prompts with the true 3D spatial context of objects. This minimizes errors caused by overlapping objects and improves the model’s understanding of scene geometry.
\end{itemize}

\section{Additional Quantitative Results}
In this section, to pursue a more comprehensive comparison, we provide additional quantitative results of the proposed \textit{\textbf{\textcolor{BurntOrange}{See}\textcolor{Emerald}{Ground}}} framework.

\subsection{Agents of Different Sizes}

\cref{tab:sup-agents} showcases the performance of different open-source VLMs on the Nr3D~\cite{achlioptas2020referit3d} validation set, evaluated across various difficulty levels and dependency types. The results highlight the compatibility of our pipeline with multiple VLM architectures, including InternVL~\cite{internvl-1, internvl-15, internvl-2} and Qwen2-VL~\cite{qwen2-vl}, across different model sizes. 

Notably, the proposed pipeline is not restricted to the specific VLMs shown in the table. It is inherently designed to be adaptable to any VLM with Optical Character Recognition (OCR) capabilities. Within our framework, OCR functionality plays a crucial role in identifying object IDs in rendered images and associating them with textual descriptions. This process enables precise alignment between 2D visual features and 3D spatial information. Consequently, the pipeline is well-suited for integration with a wide range of existing and future VLMs, further extending its applicability to 3D visual grounding tasks.

\begin{table}[t]
    \centering
    \small
    \caption{Performance comparison of different VLMs on Nr3D~\cite{achlioptas2020referit3d}.}
    \vspace{-0.2cm}
    \resizebox{\linewidth}{!}{%
    \begin{tabular}{c|cccc|c}
        \toprule
        \textbf{Agents} & \textbf{Easy} & \textbf{Hard} & \textbf{Dep.} & \textbf{Indep.} & \textbf{Overall} 
        \\
        \midrule\midrule
        InternVL2-8B & $43.6$ &  $25.8$ &  $32.6$ &  $35.4$ & $34.3$ \\
        InternVL2-26B  & $46.8$ & $29.8$  &  $34.7$ & $39.8$ &  $38.0$ \\
        Qwen2-VL-7B  & $40.8$ & $26.3$  & $31.4$ & $34.3$ & $33.3$ \\
        Qwen2-VL-72B & $\mathbf{54.5}$ & $\mathbf{38.3}$ & $\mathbf{42.3}$ & $\mathbf{48.2}$ &  $\mathbf{46.1}$ \\
        \bottomrule
    \end{tabular}%
    }
    \label{tab:sup-agents}
\end{table}



\begin{figure}[t]
    \centering
    \includegraphics[width=\linewidth]{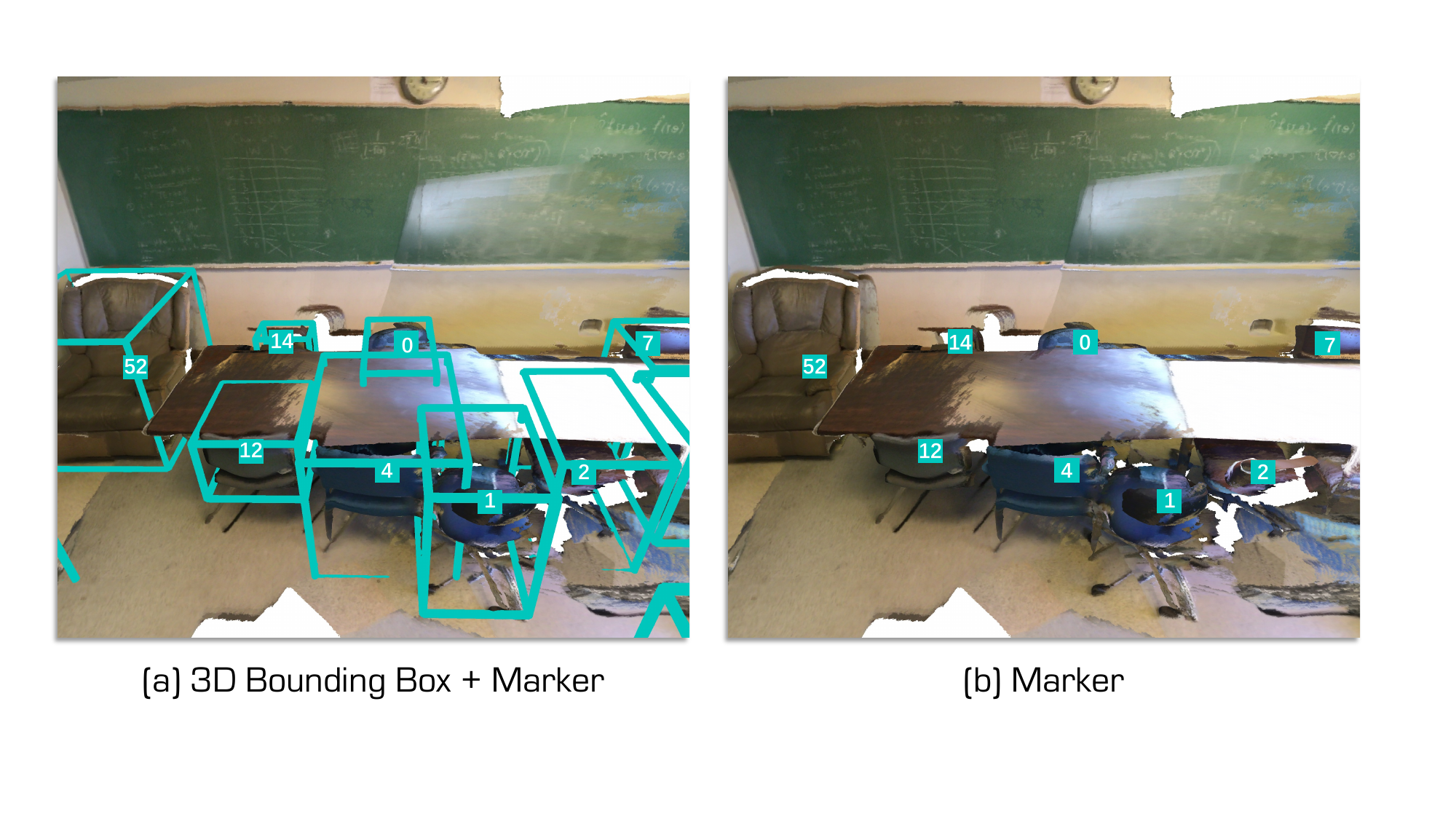}
    \caption{Illustrative examples of different visual prompts in our designs. The Marker size is enlarged for clarity.}
    \label{fig:sup-visual-prompt}
\end{figure}

\begin{table}[t]
    \centering
    \small
    \caption{Performance comparison of different visual prompts: Marker, Mask, Contour, and BBOX. Results are from $40$ randomly selected scenes out of $130$ rooms in the Nr3D \cite{achlioptas2020referit3d} validation set.}
    \vspace{-0.2cm}
    \resizebox{\linewidth}{!}{
    \begin{tabular}{c|cccc|c}
        \toprule
        \textbf{Type} & \textbf{Easy} & \textbf{Hard} & \textbf{Dep.} & \textbf{Indep.} & \textbf{Overall} 
        \\
        \midrule\midrule
        BBOX  & $53.3$ & $37.4$ &  $41.1$ &  $47.3$ &  $45.1$ \\
        Mask & $53.9$ &  $35.1$ & $39.6$ & $47.4$  & $45.0$    \\
        Contour & $56.2$ &  $37.7$ & $43.1$  &  $49.4$ & $47.5$    \\
        Marker & $\mathbf{54.8}$ & $\mathbf{39.7}$ & $\mathbf{40.2}$ & $\mathbf{51.0}$ &  $\mathbf{47.7}$ \\
        \bottomrule
    \end{tabular}%
    }
    \label{tab:sup-visual-prompt}
\end{table}

\subsection{Analysis of Visual Prompt Types}

To further explore the role of visual prompts in 3D visual grounding, we provide an analysis of alternative designs, including Mask, Contour, and BBOX, as illustrated in \cref{fig:sup-visual-prompt} and \cref{tab:sup-visual-prompt}. These visual prompts each present unique advantages and limitations, particularly when combined with 3D spatial information, as used in our method. We conducted experiments using a subset (randomly selected 40 scenes from the 130 scenes) of the Nr3D validation set. 

\begin{itemize}
    \item \noindent \textbf{Mask.} It intuitively highlights the entire object surface, making the target region explicitly visible. However, even with high transparency (as shown in \cref{fig:sup-visual-prompt} (b)), Mask can obscure surface details like texture and fine-grained patterns, which are critical for distinguishing objects. Additionally, the extra appearance information provided by Mask may be unnecessary when 3D spatial information is already available, potentially distracting the model’s attention. Moreover, generating and rendering masks for all surface points increases computational overhead, especially in complex scenes. 
    \item \noindent \textbf{BBOX.} It clearly defines spatial boundaries but introduces visual complexity due to the overlay of bounding box lines. These lines often obscure surface features (color/texture), interfering with the model's ability to interpret appearance details. In dense or overlapping object scenarios, bounding boxes can create additional confusion. Furthermore, the spatial information conveyed by BBOX prompts is redundant when 3D spatial positions are already provided, diminishing model performance.
    \item \noindent \textbf{Contour.} It represents a balance between simplicity and informativeness. By outlining object boundaries, they provide clear spatial context while avoiding the occlusion issues of Mask and BBOX. Contours also retain surface visibility, preserving critical appearance cues. The experimental results indicate that Contours perform similarly to Markers because both approaches minimize visual distractions while preserving spatial and appearance cues.
    \item \noindent \textbf{Marker.} It offers the most minimal and focused design, marking object centers without introducing visual clutter or occluding appearance features. This approach maximally preserves object details like texture and color while providing essential spatial information. The direct mapping of markers to 3D spatial positions aligns seamlessly with the 3D spatial information already used in our method, enhancing localization precision.
\end{itemize}
While Mask, Contour, and BBOX prompts each have specific strengths, their limitations -- such as visual interference or redundancy -- make Marker the most suitable choice for our framework. Its simplicity and compatibility with 3D spatial inputs ensure efficient and accurate 3D visual grounding in complex scenarios.


\subsection{Results on Different Detectors.}

\begin{table}[t]
    \centering
    \caption{Performance comparison of different 3D detectors on the ScanRefer~\cite{chen2020scanrefer} validation set. Accuracy (Acc.) is reported for each method paired with different 3D detectors.}
    \vspace{-0.15cm}
    \setlength{\tabcolsep}{15pt} 
    \begin{tabular}{l|c|c}
        \toprule
        \textbf{Method} & \textbf{3D Detector} & \textbf{Acc.} \\
        \midrule \midrule
        \multirow{2}{*}{ZSVG3D~\cite{zsvg3d}} & Mask3D~\cite{openmask3d} & $36.4$ \\
        & OVIR-3D & $19.3$ \\
        \midrule
        \multirow{3}{*}{ \textbf{\textcolor{BurntOrange}{See}\textcolor{Emerald}{Ground}}} & Ground Truth & $59.5$ \\
                                   & Mask3D~\cite{Schult23mask3d}& $44.1$ \\
                                   & OVIR-3D~\cite{ovir3d} & $30.7$ \\
        \bottomrule
    \end{tabular}
    \vspace{-0.5cm}
    \label{tab:3d_comparison}
\end{table}

\cref{tab:3d_comparison} presents a performance comparison of different 3D detectors on the ScanRefer validation set, highlighting the impact of detector choice on grounding accuracy. With the same 3D detector (Mask3D), our method significantly outperforms the previous state-of-the-art approach, ZSVG3D, achieving an accuracy of $44.1$ compared to $36.4$. 
We also explore OVIR-3D as an alternative detector. The results show that our method achieves an accuracy of $30.7$ with OVIR-3D, while ZSVG3D achieves $19.3$ under the same setting. 
Additionally, the table reveals the upper-bound performance of our method when using ground truth (GT) proposals, reaching an accuracy of $59.5$. This underscores the importance of improving 3D object detection accuracy, as better detection directly translates to enhanced grounding results. 

\subsection{Real-world Image \vs Rendered Image}


SeeGround begins with 3D object detection, which is performed directly on the 3D point cloud.
Point clouds, while sparse and noisy, inherently capture geometric details like size, shape, and spatial location. This makes the 3D detection stage less susceptible to the visual artifacts that typically affect rendered images (e.g., inconsistent lighting, color shifts).
Following this, the detected objects are used to generate rendered images from selected viewpoints. These rendered images serve as visual inputs for VLMs, combined with explicit textual descriptions and spatial information.
This workflow naturally raises questions about the impact of rendering quality on the method's performance, particularly in comparison to methods discussed in works like EmbodiedScan~\cite{wang2024embodiedscan} (Table 7), which highlight a domain gap between rendered images and real-world images.

However, unlike methods that rely purely on rendered images for learning and inference (e.g., rendering RGB images and directly training models on them), SeeGround treats rendered images as part of a multimodal input. The rendered images provide a visual representation of the scene but are supplemented by 3D spatial data, which is independent of rendering quality.  This additional spatial information reduces reliance on rendering fidelity. 

\section{More Visualization Results}


\cref{fig:supp-vis} provides additional visual examples to supplement the analysis in the main text, further illustrating the advantages of our method over previous approaches. By comparing predictions made by previous methods and Ours across various query-based 3D visual grounding tasks, we highlight the importance of appearance information (e.g., texture, color, and shape) in resolving ambiguities and improving localization accuracy. 

As shown in these examples, previous methods often fail to utilize appearance information effectively, leading to incorrect predictions when objects share similar spatial configurations or belong to the same category. For instance, queries like ``\texttt{the trash can next to the blackboard}'' or ``\texttt{the monitor in front of the black keyboard}'' require fine-grained differentiation based on appearance attributes. Previous methods tend to misidentify nearby or visually similar objects due to their limited ability to integrate these attributes into the grounding process. In contrast, our method incorporates appearance information explicitly through depth-aware visual prompting, enabling more accurate alignment of textual descriptions with 3D spatial and visual cues.

These supplementary results emphasize the critical role of appearance information in 3D visual grounding and demonstrate how our method effectively leverages this information to address ambiguities. By incorporating visual features alongside spatial reasoning, our approach achieves significant improvements in challenging scenarios, further validating the findings presented in the main text.

\section{Broader Impact \& Limitations}
\label{sec:limitation}
In this section, we elaborate on the broader impact and potential limitations of this work.

\subsection{Broader Impact}
Our approach bridges 3D data and 2D VLMs, making 3D visual grounding accessible in zero-shot settings. This design reduces reliance on large-scale 3D-specific datasets and annotations, enabling scalable deployment. By focusing on integrating 2D rendered images with spatial descriptions, our method highlights the importance of appearance features like color, texture, and orientation, which are often overlooked in previous zero-shot approaches. Applications range from assistive technologies to robotics and augmented reality, where robust object localization can enhance usability and accessibility.
Moreover, the use of visual prompts, especially the Marker-based design, introduces an interpretable mechanism for aligning visual and spatial information. This improves transparency and trust in AI systems, allowing stakeholders to better understand the reasoning behind model predictions. 

\subsection{Potential Limitations}
Despite its advancements, our method has some limitations. It relies on accurate 3D object detection and spatial data, making it vulnerable to errors in preprocessing. Misaligned bounding boxes or missing objects can propagate through the pipeline, reducing localization accuracy. Marker-based visual prompts, while simple and effective, may struggle in cluttered scenes requiring richer contextual information and can obscure very small objects, complicating precise localization.
The method leverages 2D-3D alignment without requiring highly accurate rendered images, but consistent alignment remains crucial for effective multimodal fusion. Significant deviations in rendered images -- caused by inaccurate camera parameters or low-quality point clouds -- can compromise alignment between 2D prompts and 3D spatial descriptions. This issue is exacerbated in cluttered/dynamic scenes, where rendering delays can lead to mismatches between 2D prompts and real-time 3D data, causing errors in grounding. For instance, in scenes with moving objects, outdated rendered views may misrepresent object positions, leading to incorrect target identification.
Future work could enhance multimodal alignment robustness under noisy or sparse data, improve real-time efficiency, and better handle dynamic and cluttered environments, broadening the method’s applicability to complex real-world scenarios.
\section{Public Resource Used}
\label{sec:ack}
In this section, we acknowledge the use of the following public resources, during the course of this work:

\begin{itemize}
    \item Pytorch \footnote{\url{https://github.com/pytorch/pytorch}} \dotfill Pytorch License 
    \item Pytorch3D \footnote{\url{https://github.com/facebookresearch/pytorch3d}} \dotfill BSD-Style License 
    \item Open3D \footnote{\url{https://github.com/isl-org/Open3D}} \dotfill MIT license
    \item ScanRefer\footnote{\url{https://daveredrum.github.io/ScanRefer/}} \dotfill ScanRefer License
    \item Nr3D \footnote{\url{https://github.com/referit3d/referit3d}} \dotfill MIT License
    \item Qwen2-VL\footnote{\url{https://github.com/QwenLM/Qwen2-VL}} \dotfill Apache License 2.0
    \item InternVL2 \footnote{\url{https://huggingface.co/OpenGVLab/InternVL2-26B}} \dotfill MIT license
    \item ZSVG3D \footnote{\url{https://github.com/CurryYuan/ZSVG3D}} \dotfill Unknown
    \item vLLM \footnote{\url{https://github.com/vllm-project/vllm}} \dotfill Apache License 2.0
    \item OpenScene \footnote{\url{https://github.com/pengsongyou/openscene}} \dotfill Apache License 2.0
    \item vil3dref \footnote{\url{https://github.com/cshizhe/vil3dref}} \dotfill Unknown
    \item OpenIns3D \footnote{\url{https://github.com/Pointcept/OpenIns3D}} \dotfill MIT License
    \item EmboddiedScan \footnote{\url{https://github.com/OpenRobotLab/EmbodiedScan}}  \dotfill Apache License 2.0
    \item LAR \footnote{\url{https://github.com/eslambakr/LAR-Look-Around-and-Refer}}  \dotfill MIT License
\end{itemize}

\clearpage\clearpage
\begin{figure*}[t]
\centering
\includegraphics[width=0.9\linewidth]{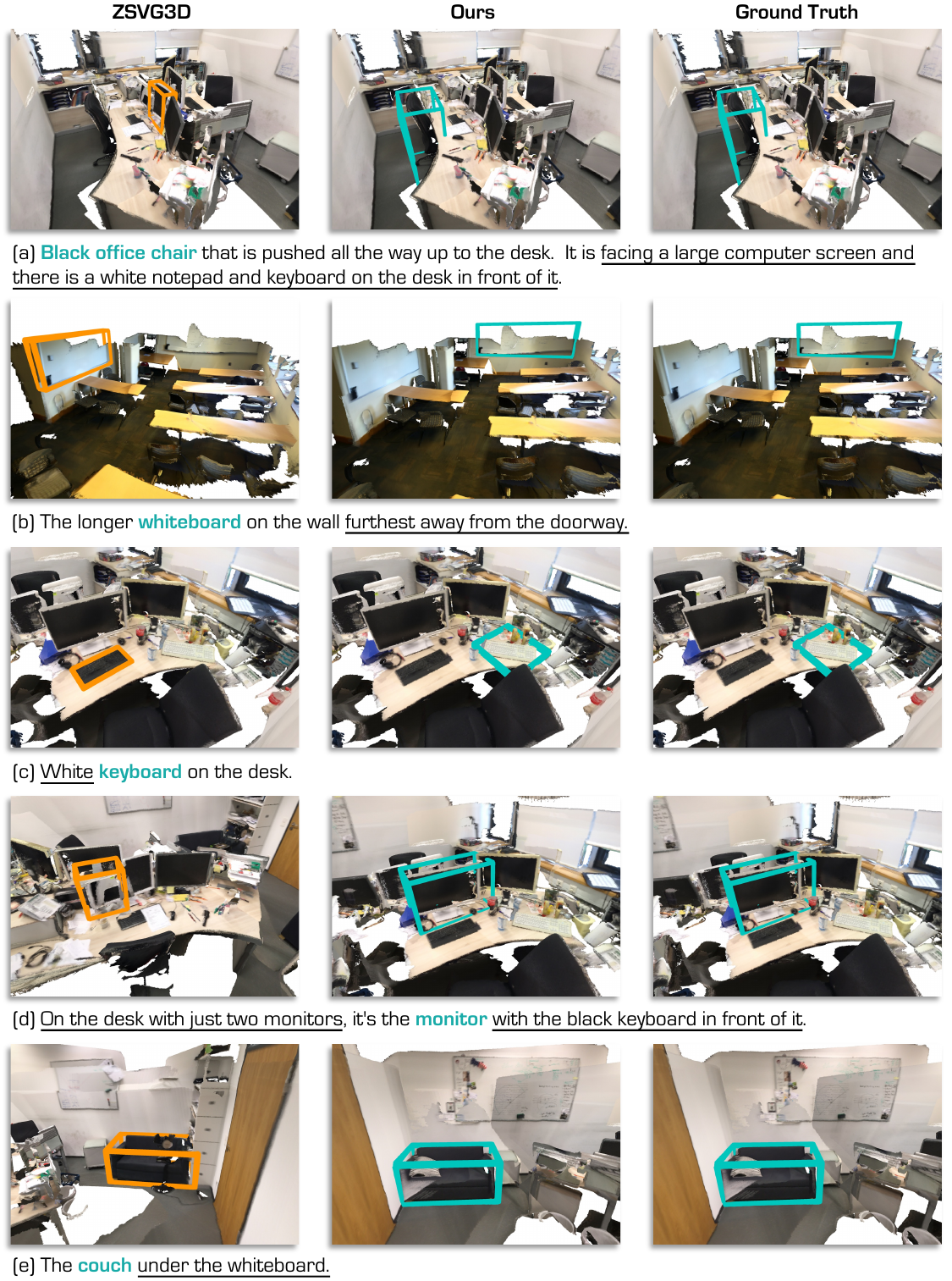}
\end{figure*}

\begin{figure*}[ht]
\centering
\includegraphics[width=0.9\linewidth]{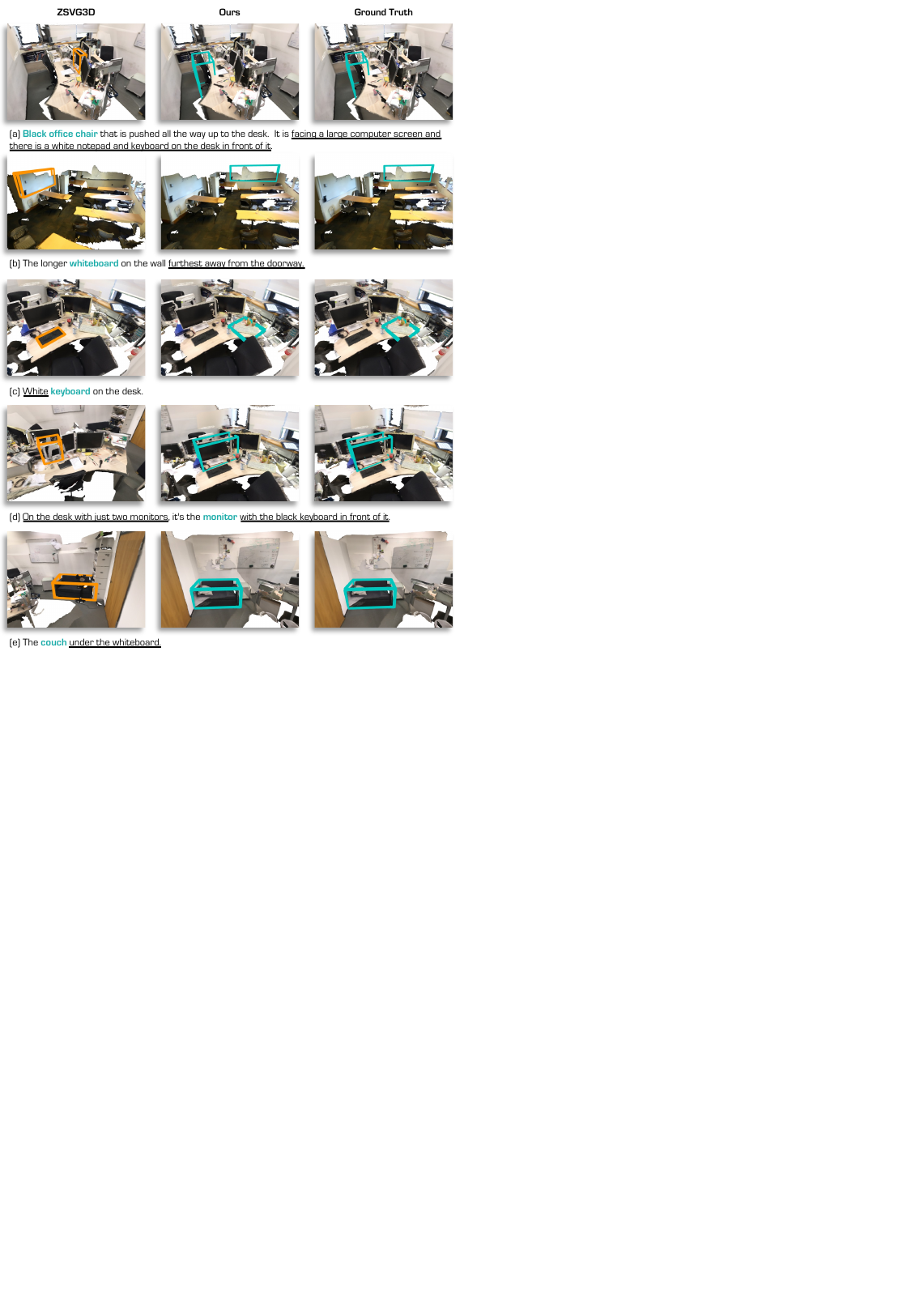}
\caption{Illustration of SeeGround's capability to resolve ambiguities in the 3D visual grounding task. Incorrectly identified objects ({\color{BurntOrange}{Orange}}) and correctly identified objects ({\color{Emerald}{Green}}) are indicated to differentiate prediction accuracy, \underline{key cues} are underlined.}
\label{fig:supp-vis}
\end{figure*}

\clearpage\clearpage
{
    \small
    \bibliographystyle{ieeenat_fullname}
    \bibliography{main}
}

\end{document}